%% file: main.tex
\documentclass[nohyperref]{article}

\usepackage{microtype}
\usepackage{graphicx}
\usepackage{booktabs} 

\usepackage[pagebackref=true]{hyperref}

\usepackage[accepted]{icml2023}

\usepackage{amsmath}
\usepackage{amssymb}
\usepackage{mathtools}
\usepackage{amsthm}
\usepackage{bbold}
\usepackage[normalem]{ulem}
\usepackage{caption}
\usepackage{subcaption}
\usepackage{centernot}
\usepackage{soul}
\usepackage{multirow}
\usepackage{comment}

\usepackage[capitalize,noabbrev]{cleveref}

\usepackage{amsfonts}

\usepackage[textsize=tiny]{todonotes}

\usepackage{thm-restate}
\usepackage{xcolor}

\usepackage[utf8]{inputenc}
\usepackage{amsmath}
\usepackage{amsthm}
\usepackage{amssymb}
\usepackage{cleveref}

\usepackage{natbib}

\definecolor{RoyalBlue}{rgb}{0,0,0.8}
\hypersetup{colorlinks, citecolor=RoyalBlue, linkcolor=RoyalBlue}

\newcommand{\q}[1]{\textcolor{red}{#1}}

\newcommand{\sd}[1]{{{\footnotesize±}{\scriptsize#1}}}

\declaretheorem[name=Lemma,numberwithin=section]{lemma}

\declaretheorem[name=Definition,numberwithin=section]{definition}

\icmltitlerunning{Local Vertex Colouring Graph Neural Networks}

\newcommand*{\model}{SGN}
\newcommand*{\citeseer}{{\sc CiteSeer}}
\newcommand*{\cora}{{\sc Cora}}
\newcommand*{\pubmed}{{\sc Pubmed}}
\newcommand*{\cornell}{{\sc Cornell}}
\newcommand*{\texas}{{\sc Texas}}
\newcommand*{\wisconsin}{{\sc Wisconsin}}
\newcommand*{\chameleon}{{\sc Chameleon}}
\newcommand*{\squirrel}{{\sc Squirrel}}
\newcommand*{\amazonphoto}{{\sc Photo}}
\newcommand*{\amazoncomputers}{{\sc Computers}}
\newcommand*{\dd}{{\sc D\&D}}
\newcommand*{\nci}{{\sc NCI1}}
\newcommand*{\proteins}{{\sc PROTEINS}}
\newcommand*{\enzymes}{{\sc ENZYMES}}
\newcommand*{\imdbb}{{\sc IMDB-BINARY}}
\newcommand*{\baseline}{{\sc Baseline}}

\begin{document}

\twocolumn[
\icmltitle{Local Vertex Colouring Graph Neural Networks}
\icmlsetsymbol{equal}{*}

\begin{icmlauthorlist}
\icmlauthor{Shouheng Li}{anu,csiro}
\icmlauthor{Dongwoo Kim}{postech}
\icmlauthor{Qing Wang}{anu}
\end{icmlauthorlist}

\icmlaffiliation{anu}{School of Computing, Australian National University, Canberra, Australia}
\icmlaffiliation{postech}{CSE \& GSAI, POSTECH, Pohang, South Korea}
\icmlaffiliation{csiro}{Data61, CSIRO, Canberra, Australia}

\icmlcorrespondingauthor{Dongwoo Kim}{dongwoo.kim@postech.ac.kr}
\icmlcorrespondingauthor{Qing Wang}{qing.wang@anu.edu.au}

\icmlkeywords{Machine Learning, ICML}

\vskip 0.3in
]

\printAffiliationsAndNotice{}

\input{sections/abstract.tex}
\input{sections/intro.tex}
\input{sections/related_work.tex}
\input{sections/preliminary.tex}
\input{sections/method.tex}

\input{sections/model.tex}

\input{sections/experiment.tex}
\input{sections/conclusion.tex}

\subsection*{Acknowledgements}
We thank Professor Brendan Mckay for helpful suggestions. We acknowledge the support of the Australian Research Council under Discovery Project DP210102273. This work was also partly supported by Institute of Information \& communications Technology Planning \& Evaluation (IITP) grant funded by the Korea government (MSIT) (No.2019-0-01906, Artificial Intelligence Graduate School Program(POSTECH)) and National Research Foundation of Korea (NRF) grant funded by the Korea government (MSIT) (NRF-2021R1C1C1011375).

\bibliography{references}
\bibliographystyle{icml2023}

\newpage
\appendix
\onecolumn
\input{sections/appendix}

\end{document}

%% file: sections/abstract.tex
\begin{abstract}
In recent years, there has been a significant amount of research focused on expanding the expressivity of Graph Neural Networks (GNNs) beyond the Weisfeiler-Lehman (1-WL) framework. 
While many of these studies have yielded advancements in expressivity, they have frequently come at the expense of decreased efficiency or have been restricted to specific types of graphs.
In this study, we investigate the expressivity of GNNs from the perspective of graph search. Specifically, we propose a new vertex colouring scheme and demonstrate that classical search algorithms can efficiently compute graph representations that extend beyond the 1-WL. 
We show the colouring scheme inherits useful properties from graph search that can help solve problems like graph biconnectivity.
Furthermore, we show that under certain conditions, the expressivity of GNNs increases hierarchically with the radius of the search neighbourhood.
To further investigate the proposed scheme, we develop a new type of GNN based on two search strategies, breadth-first search and depth-first search, highlighting the graph properties they can capture on top of 1-WL. Our code is available at \url{https://github.com/seanli3/lvc}.
\end{abstract}

%% file: sections/intro.tex
\section{Introduction}
Graph neural networks (GNNs) have emerged as the de-facto method for representation learning on graphs. One popular architecture of GNNs is the message-passing neural networks (MPNNs) which propagate information between vertices along edges~\citep{gilmer2017neural, kipf2016semi,velivckovic2017graph}.  In~\citet{xu2018powerful}, it is shown that the design of MPNNs aligns with the Weisfeiler-Lehman (1-WL) test, a classical algorithm for testing graph isomorphism. Therefore, the expressivity of MPNNs is upper-bounded by the 1-WL test. Intuitively, if two vertices have the same computational/message-passing graph in an MPNN, they are indistinguishable.

Recent studies attempted to increase the expressivity of GNNs beyond the 1-WL. One direction is to extend GNNs to match higher-order WL tests~\citep{morris2019weisfeiler, morris2020weisfeiler, maron2019provably, geerts2022expressiveness}. While these methods offer improved expressivity, they come at the cost of decreased efficiency, as higher-order WL tests are known to be computationally expensive.
Another line of research focuses on incorporating graph substructures into feature aggregation~\citep{bodnar2021weisfeiler, bodnar2021weisfeilercellular}. However, these approaches often rely on task-specific, hand-picked substructures. One other strategy is to enhance vertex or edge features with additional distance information relative to target vertices~\citep{YouYL19positionaware,LiWWL20distanceencode}.
Despite the efforts, \citet{anonymous2023rethinking} have shown that MPNNs cannot solve the biconnectivity problem, which can be efficiently solved using the depth-first search algorithm~\citep{Tarjan1974-eb}. 

In light of the aforementioned understanding, one may question how the design of GNNs can surpass the limitations of 1-WL to address issues that cannot be resolved by MPNNs. 
In this work, we systematically study an alternative approach to MPNN, which propagates information along graph search trees. This paper makes the following contributions:
\begin{itemize}
    \item We design a novel colouring scheme, called \emph{local vertex colouring} (LVC), based on breath-first and depth-first search algorithms, that goes beyond 1-WL.
    \item We show that LVC can learn representations to distinguish several graph properties such as biconnectivity, cycles, cut vertices and edges, and \textit{ego short-path graphs} (ESPGs) that 1-WL and MPNNs cannot.
    \item We analyse the expressivity of LVC in terms of  \emph{breadth-first colouring} and \emph{depth-first colouring}, and provide systematical comparisons with 1-WL and 3-WL. 
    \item We further design a graph search-guided GNN architecture, \textit{Search-guided Graph Neural Network}~(\model{}), which inherits the properties of LVC.
\end{itemize}

%% file: sections/related_work.tex
\section{Related Work}
\paragraph{Graph isomorphism and colour refinement.} 
The graph isomorphism problem concerns whether two graphs are identical topologically, e.g. for two graphs $G$ and $H$, whether there is a bijection $f:V_G \rightarrow V_H$ such that any two vertices $u$ and $v$ are adjacent in $G$ if and only if $f(u)$ and $f(v)$ are adjacent in $H$. If such a bijection exists, we say $G$ and $H$ are \emph{isomorphic} ($G\simeq H$).

Let $C$ be a set of colours. A \emph{vertex colouring refinement} function $\lambda: V \rightarrow C$ assigns each vertex $v\in V$ with a colour $\lambda(v)\in C$. This assignment is performed iteratively until the vertex colours no longer change. Colour refinement can be used to test graph isomorphism by comparing the multisets of vertex colours $\{\!\!\{\lambda(v): v\in V_G\}\!\!\}$ and $\{\!\!\{\lambda(u): u\in V_H\}\!\!\}$, given that $\lambda(\cdot)$ is invariant under isomorphic permutations. A classic example of such tests is the \emph{Weisfeiler-Lehman (WL) test} ~\citep{weisfeiler1968reduction}, which assigns a colour to a vertex based on the colours of its neighbours.
\citet{cai1992optimal} extends 1-WL to compute a colour on each k-tuple of vertices; this extension is known as the \emph{k-dimensional Folklore Weisfeiler-Lehman algorithms} (k-FWL).
We may apply a colouring refinement to all vertices in $G$ iteratively until vertex colours are stabilised. 

Let $(G,\lambda^i)$ denote a colouring on vertices of $G$ after applying a colour refinement function $i$ times, i.e., after the $i$-th iteration, and $P(\lambda^i)$ a partition of the vertex set induced by the colouring $(G,\lambda^i)$. For two vertex partitions $P(\lambda^i)$ and $P(\lambda^j)$ on $G$, if every element
of $P(\lambda^i)$ is a (not necessarily proper) subset of an element of $P(\lambda^j)$, we say $P(\lambda^i)$ \emph{refines} $P(\lambda^j)$. 
When $P(\lambda^{j})\equiv P(\lambda^{j+1})$, we call $\lambda^j$ a \emph{stable colouring} and $P(\lambda^{j})$ a \emph{stable partition} of $G$.

\paragraph{GNNs beyond 1-WL.}
MPNN is a widely adopted graph representation learning approach in many applications. However, there are a few caveats. Firstly, as shown by \citet{xu2018powerful}, the expressive power of MPNN is upper-bounded by 1-WL, which is known to have limited power in distinguishing isomorphic graphs. Secondly, to increase the receptive field, MPNN needs to be stacked deeply, which causes over-smoothing~\citep{zhao2019pairnorm,chen2020measuring} and over-squashing~\citep{topping22oversquashing} that degrade performance. Recent research aims to design more powerful GNNs by incorporating higher-order neighbourhoods~\citep{maron2019provably,morris2019weisfeiler}. However, these methods incur high computational costs and thus are not feasible for large datasets. Other methods alter the MPNN framework or introduce extra heuristics to improve expressivity~\citep{bouritsas2022improving,bodnar2021weisfeiler,bodnar2021weisfeilercellular,bevilacqua2021equivariant,wijesinghe2021new}. However, while these methods are shown to be more powerful than 1-WL, it is still unclear what additional properties they can capture beyond 1-WL.

\paragraph{GNNs in learning graph algorithms.}
\citet{Velickovic_neuralexecutoin} show that MPNN can imitate classical graph algorithms to learn shortest paths (Bellman-Ford algorithm) and minimum spanning trees (Prim’s algorithm). 
\citet{dobrik_neuralbipartite} show that MPNNs can execute the more complex Ford-Fulkerson algorithm, which consists of several composable subroutines, for finding maximum flow. 
\citet{loukas2020graph} further shows that certain GNN can solve graph problems like cycle detection and minimum cut, but only when its depth and width reach a certain level.
\citet{xu20_gnnreasoning} show that the ability of MPNN to imitate complex graph algorithms is restrained by the alignment between its computation structure and the algorithmic structure of the relevant reasoning process. 
One such example, as shown by \citet{anonymous2023rethinking}, is that MPNN cannot solve the biconnectivity problem, despite that this problem has an efficient algorithmic solution linear to graph size. 

%% file: sections/preliminary.tex
\section{Preliminaries}
\label{sec:pre}
Let $\{\cdot\}$ denote sets and $\{\!\!\{\cdot\}\!\!\}$  multisets. 
We consider undirected simple graphs $G = (V, E)$ where $V$ is the vertex set and $E$ is the edge set. 
We use $\lvert\cdot\rvert$ to denote the cardinality of a set/multiset/sequence, $e_{vu} = \{v,u\}$ an undirected edge connecting vertices $v$ and $u$, and $\vec{e}_{vu} = (v,u)$ a directed edge that starts from vertex $v$ and ends at vertex $u$. Thus, $e_{vu} = e_{uv}$ and $\vec{e}_{vu} \neq \vec{e}_{uv}$. We use $d(v,u)$ to denote the shortest-path distance between vertices $v$ and $u$.
A \emph{$\delta$-neighborhood} of a vertex $v$ is a set of vertices within the distance $\delta$ from $v$, i.e. $N_{\delta}(v)=\{u\in V:1\leq d(v,u)\leq \delta\}$.
A path $\mathcal{P}_{w_0w_k}$ of length $k$ in $G$, called \emph{k-path}, is a sequence $(w_0, w_1, ..., w_k)$ of distinct vertices such that $(w_{i-1}, w_{i}) \in E$ for $i=1,2,...,k$. 

\begin{figure}[t]
\centering
\begin{subfigure}{.3\columnwidth}
\captionsetup{width=.3\columnwidth}
        \includegraphics[clip,width=\textwidth]{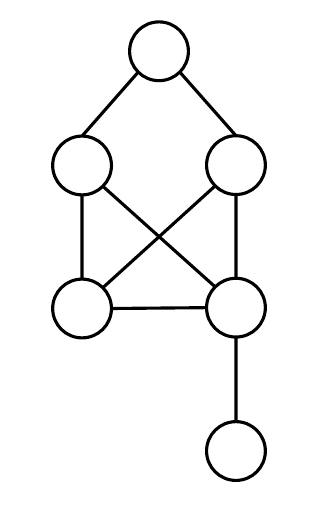}
\subcaption{}
\label{subfig:edge_type_a}
\end{subfigure}%
\hfill
\begin{subfigure}{.3\columnwidth}
\captionsetup{width=.9\linewidth}
        \includegraphics[clip,width=\textwidth]{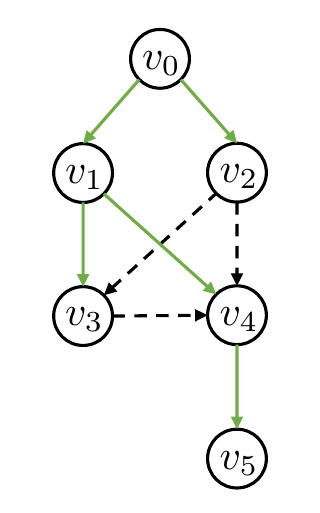}
\subcaption{BFS}
\label{subfig:edge_type_b}
\end{subfigure}
\hfill
\begin{subfigure}{.3\columnwidth}
\captionsetup{width=.9\linewidth}
        \includegraphics[clip,width=\textwidth]{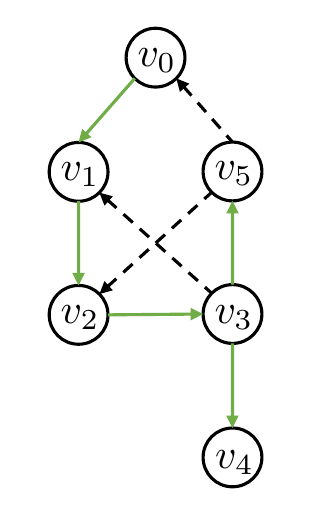}
\subcaption{DFS}
\label{subfig:edge_type_c}
\end{subfigure}
\caption{Tree edges (green solid lines) and back edges (black dashed lines) classified by BFS and DFS. The subscripted labels $v_0$, \dots, $v_5$ denote the visit sequence of each vertex, e.g. $v_1$ is visited after $v_0$.}
\label{fig:edge_type}
\end{figure}

\paragraph{Graph searching.} 
Graph traversal, or \emph{graph searching}, visits each vertex in a graph. \emph{Breadth-First Search} (BFS) and \emph{Depth-First Search} (DFS) are the two most widely used graph search algorithms, which differ in the order of visiting vertices. Both methods start at a vertex $v$. BFS first visits the direct neighbours in $N_1(v)$ of $v$ and then the neighbours in $N_2(v)$ that have not been visited, etc. The idea is to process all vertices in $N_i$ of distance (or level) $i$ from $v$ before processing vertices at level $i+1$ or greater. The process is repeated until all vertices reachable from $v$ have been visited. In DFS we instead go as ``deep" as possible from vertex $v$ until no new vertices can be visited. Then we backtrack and try other neighbours that were missed from the farthest in the search paths until all vertices are visited. The visited vertices and the edges along the search paths form a BFS/DFS search tree. For example, the solid lines in \cref{subfig:edge_type_b,subfig:edge_type_c} represent two different search trees for the graph in \cref{subfig:edge_type_a}. Given a graph, generally, there are many ways to construct different search trees by selecting different starting points and different edges to visit vertices.

\paragraph{Visiting order subscripting.} Vertices being visited in a graph search form a linear order. We use subscripts to label this order based on their discovery/first-visited time~\citep{Cormen_algointro}. Given a search tree, we say $v_i$ \emph{precedes} $v_j$, denoted as $v_i\prec v_j$, if $v_i$ is discovered before $v_j$. $i$ and $j$ are subscripts that indicate the relation: for all $i$ and $j$, $v_i\prec v_j$ if and only if $i< j$. Each vertex is discovered once, so a subscript ranges from 0 to $N-1$ in a graph $G$. $v_0$ is called the root. \cref{fig:edge_type} shows how the subscripting is used to indicate vertex visiting orders. 


\paragraph{Tree and back edges.} For an undirected simple graph $G=(V,E)$, BFS/DFS categorise the edges in $E$ into \emph{tree edges} and \emph{back edges}. 

Tree edges and back edges are both directed, i.e., $(v,u)\neq (u,v)$. Let $T_v$ denote a search tree rooted at vertex $v$. A tree edge is an edge in the search tree $T_v$, starting from an early-visited vertex to a later-visited vertex. A back edge starts from a later-visited vertex to an early-visited vertex.
A directed edge $(v_i, v_j)$ is either a tree edge (if $i<j$), or a back edge (if $i>j$).
For example, dashed lines in \cref{subfig:edge_type_b,subfig:edge_type_c} represent back edges in BFS and DFS search trees, respectively. 
In BFS, a back edge $(v_i, v_j)$ connects vertices at the same or adjacent level, i.e. $d(v_0,v_i) =d(v_0,v_j)$ or $|d(v_0,v_i) - d(v_0,v_j)| = 1$~\citep{Cormen_algointro}. For this reason back edges in BFS are sometimes referred to as \emph{cross edges}.
In DFS, a back edge $(v_i,v_j)$ connects a vertex $v_i$ and its non-parent ancestor $v_j$; therefore we always have $v_j\prec v_i$ (or $i>j$). 

We use $E_{\text{tree}}^{T_v}$ and $E_{\text{back}}^{T_v}$ to denote the tree edge set and the back edge set with respect to a search tree $T_v$ rooted at $v$. 





%% file: sections/method.tex
\section{Local Vertex Colouring}

In this section, we introduce the building blocks of a search-guided colouring scheme and demonstrate its expressive power in distinguishing non-isomorphic graphs. We also discuss how this search-guided colouring scheme can solve the biconnectivity and ego shortest-path graph problems that cannot be solved by MPNN. 

\subsection{Search-guided Vertex Colour Update}
We first describe a way to update vertex colours guided by graph searching. As described in \cref{sec:pre}, graph searching w.r.t. a search tree $T_v$ categorises edges into two sets: a tree edge set $E_{\text{tree}}^{T_v}$ and a back edge set $E_{\text{back}}^{T_v}$. 
Instead of propagating messages along all edges like MPNN, we design a scheme that propagates vertex information, i.e., vertex colours, along tree edges and back edges.  Given a search tree $T_v$, we define the vertex colouring refinement function on each vertex $u$ as follows
\begin{equation}
\label{eqn:lvc-all}
    \lambda^{l+1}(u):= \rho(\lambda^{l}(u), \{\!\!\{\lambda_v^{l+1}(u): v\in N_{\delta}(u)\}\!\!\})
\end{equation}
where $\rho(\cdot)$ is an injective function and $\lambda_v^{l+1}(u)$ is the vertex colour computed, based on the search tree $T_v$, as  
\begin{align}
\label{eqn:lvc}
\begin{split}
    &\lambda_v^{l+1}(u):=\\
    &\phi\left(\lambda^{l}(u), \psi\left(\{\!\!\{\lambda_v^{l}(w): w \in \eta(u, E_{\text{tree}}^{T_v}, E_{\text{back}}^{T_v})\}\!\!\}\right)\right)
\end{split}
\end{align}

where $\phi(\cdot)$ and $\psi(\cdot)$ are injective functions, 
and $l$ is the number of the current iteration. 
Let $\mathbb{P}(E)$ and $\mathbb{P}(V)$ be the power sets of $E$ and $V$, respectively. The function $\eta: V \times \mathbb{P}(E) \times \mathbb{P}(E) \rightarrow \mathbb{P}(V)$ takes a vertex $u\in V$ to be coloured, a tree edge set  $E_{\text{tree}}^{T_v}\in \mathbb{P}(E)$, and a back edge set $E_{\text{back}}^{T_v}\in \mathbb{P}(E)$ as input, and produces a vertex set in $\mathbb{P}(V)$.

At the first iteration, $\lambda^0_v(w)$ and $\lambda^0(w)$ are the initial colours of $w$. There are two steps in the colouring scheme. The first step is \emph{search-guided colour propagation}
(\cref{eqn:lvc}), where vertex colours are propagated along the search paths of each $T_v$ based on $\eta(u, E_{\text{tree}}^{T_v}, E_{\text{back}}^{T_v})$. After this step, a vertex $u$ obtains a colour $\lambda^{l+1}_v(u)$ w.r.t. each root vertex $v\in N_{\delta}(u)$. We obtain in total $|N_{\delta}(u)|$ colours for $u$. The second step is \emph{neighbourhood aggregation} (\cref{eqn:lvc-all}), where the $|N_{\delta}(u)|$ colours obtained from the first step are aggregated and used to compute a new colour $\lambda^{l+1}(u)$ for $u$. These two steps are repeated with the new vertex colours.
 
When $N_\delta(v) \neq V$, we call this vertex colouring scheme \emph{$\delta$-local vertex colouring} (LVC-$\delta$).
LVC-$\delta$ is applied to colour vertices in $G$ iteratively until vertex colours are stabilised.
We omit superscript and use $\lambda$ to refer to a stable colouring.




\paragraph{Search order permutation.}
Graph searching may encounter cases where a search algorithm needs to decide a priority between two or more unvisited vertices (a tie). In such cases, the search algorithm can visit any vertex in the tie. This yields different vertex visiting orders, we call this permutation in vertex visiting order \textit{search order permutation}. 
For example, in \cref{subfig:edge_type_b}, the BFS rooted at vertex $v_0$ faces a tie, where it can visit either $v_1$ or $v_2$ first since both $v_1$ and $v_2$ are adjacent to $v_0$. \cref{subfig:edge_type_b} shows the search trajectory when visiting $v_1$ first; however, if the BFS visits $v_2$ first, then tree edges and back edges will be different. 

\paragraph{Design considerations.}
A key function that controls the colouring in \autoref{eqn:lvc} is $\eta$. For instance, we can make LVC-$\delta$ identical to 1-WL, if we define $\eta(u, E_{\text{tree}}^{T_v}, E_{\text{back}}^{T_v}) = \{w:(u,w) \in E_{\text{tree}}^{T_v}\vee (w,u) \in E_{\text{tree}}^{T_v}\vee (u,w) \in E_{\text{back}}^{T_v} \vee (w,u) \in E_{\text{back}}^{T_v}\}$. We name it \emph{1-WL-equivalent LVC}. In this definition, $\eta$ effectively yields all neighbouring vertices of $u$, making it equivalent to 1-WL. 

Since the design of $\eta$ is crucial, we hereby introduce two key points that should be considered when designing $\eta$. 
\begin{itemize}
    \item $\eta$ should be invariant to search order permutation, i.e. a change of vertex visiting order should not alter the output of $\eta$. If $\eta$ is not invariant to search order permutation, the colouring scheme will not be permutation invariant. The 1-WL equivalent LVC example in the previous paragraph is invariant to search order permutation. 
    \item $\eta$ should inherit properties of a search algorithm that are informative about identifying graph structure. Graph searches like BFS and DFS are widely used in graph algorithms to capture structural properties such as cycles and biconnectivity. $\eta$ should be designed to incorporate such structural information in vertex colours.
\end{itemize}


\begin{figure}[t!]
\centering
\begin{subfigure}{.44\columnwidth}
\includegraphics[clip,width=\textwidth]{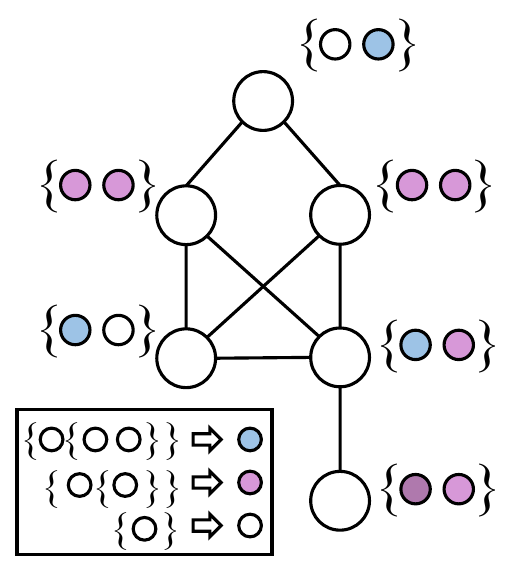}
\subcaption{}
\label{subfig:bfc1}
\end{subfigure}%
\hfill
\begin{subfigure}{.25\columnwidth}
\includegraphics[clip,width=\textwidth]{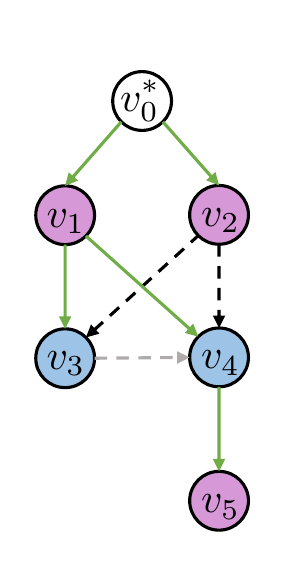}
\subcaption{}
\label{subfig:bfc2}
\end{subfigure}
\hfill
\begin{subfigure}{.25\columnwidth}
\includegraphics[clip,width=\textwidth]{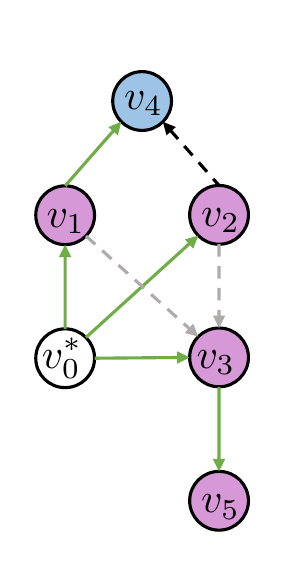}
\subcaption{}
\label{subfig:bfc3}
\end{subfigure}
\caption{
A graph and its vertex colours after the first BFC iteration. \ref{subfig:bfc1} shows the uncoloured graph, where the vertex colours obtained from \ref{subfig:bfc2} and  \ref{subfig:bfc3} are shown next to each vertex. 
\ref{subfig:bfc2} and \ref{subfig:bfc3} show BFC with two different roots (marked with *), respectively, where each vertex is assigned a new colour by BFC. The colour map is shown in the bottom left corner. The subscripted labels $v_0$, \dots, $v_5$ denote the visit sequence of each vertex, e.g. $v_1$ is visited after $v_0$.
For brevity, we show BFC with only two roots.} 
\label{fig:bfc}
\end{figure}

\subsection{BFS-guided Colouring}
BFS is perhaps the most widely used graph search algorithm and its applications include finding shortest paths, minimum spanning tree, cycle detection, and bipartite graph test. We design $\eta$ for BFS, denoted as $\eta_{\text{bfc}}$, as 
\begin{align}
\label{eqn:sigma_bf}
    &\eta_{\text{bfc}}(u, E_{\text{tree}}^{T_v}, E_{\text{back}}^{T_v}) =& \\
     &\left\{o: (o,u) \in E_{\text{tree}}^{T_v} \vee 
    \left((o, u) \in E_{\text{back}}^{T_v} \wedge
    d(v,o) \neq d(v,u)\right)
    \right\}& \notag
\end{align}
The part $\{o: (o,u) \in E_{\text{tree}}^{T_v}\}$ preserves vertices that lead to vertex $u$ via tree edges. $\{o: (o, u) \in E_{\text{back}}^{T_v}\}$ preserves vertices that lead to vertex $u$ via back edges. The condition $d(v,o) \neq d(v,u)$ ensures only back edges connecting vertices at different levels are included.
The vertex colouring scheme using $\eta_{\text{bfc}}$ is called \textit{breadth-first colouring} (BFC).

\cref{fig:bfc} shows BFC rooted at two different vertices, where the grey dashed lines indicate back edges that are excluded by BFC. In \cref{subfig:bfc2}, when $u = v_4$, $\eta_{\text{bfc}}$ returns $v_1$ and $v_2$, but excludes $v_3$ because it is at the same level as $v_4$. 

An important property of BFS is that tree edges form the shortest paths between a root to other vertices, e.g., in \cref{subfig:bfc2} $(v_0,v_1)$ and $(v_1,v_4)$ form the shortest path $(v_0,v_1,v_4)$ between $v_0$ and $v_4$. 
There are two categories of back edges in a BFS tree: the ones connecting vertices across two adjacent levels and the ones connecting vertices at the same level~\citep{Cormen_algointro}. The first category of back edges forms alternative shortest paths, e.g., $(v_0,v_2)$ and $(v_2,v_4)$ form another shortest path $(v_0,v_2,v_4)$ between $v_0$ and $v_4$. The second category of back edges does not participate in any shortest path. 
$\eta_{\text{bfc}}$ only preserves the first category of back edges and excludes the second category by the condition $d(v,o) \neq d(v,u)$. 
All shortest paths from a root to a vertex, e.g., $(v_0,v_1, v_4)$ and $(v_0,v_2,v_4)$, form an induced shortest path graph (SPG)~\citep{Wang2021-shortestpathgraph}, which is permutation invariant. For example, if we swap the search order between $v_1$ and $v_2$ in \cref{subfig:bfc2}, $(v_1,v_4)$ becomes a back edge and $(v_2,v_4)$ becomes a tree edge, but $\eta_{\text{bfc}}$ returns the same vertex set. Hence, BFC defined by Equations \ref{eqn:lvc-all}, \ref{eqn:lvc}, and \ref{eqn:sigma_bf} is also permutation invariant. 

BFC is referred as BFC-$\delta$ if the search range is limited to a $\delta$-hop neighbourhood for each root vertex.

\paragraph{Distinguishing shortest-path graphs.} 
\citet{Velickovic_neuralexecutoin} show that MPNN can imitate classical graph algorithms to learn shortest paths (Bellman-Ford algorithm) and minimum spanning trees (Prim’s algorithm). \citet{xu20_gnnreasoning} further show that MPNN is theoretically suitable for learning tasks that are solvable using dynamic programming. Since the base form of BFC (i.e. BFC-1) aligns with MPNN (will be discussed later), BFC also inherits these properties. However, we are more interested in tasks which BFC can do but MPNN cannot. We show that one of such tasks is to distinguish ego shortest-path graphs. 

\begin{definition}
For two vertices $v, u\in V_G$, a \emph{shortest-path graph} $SPG(v,u)$ is a subgraph of $G$, where $SPG(v,u)$ contains all and only vertices and edges occurring in the shortest paths between $u$ and $v$.
\end{definition}

\begin{restatable}[]{lemma}{bfcspg}
\label{lemma:bfc2spg}
Let $(u,v)$ and $(u',v')$ be two pairs of vertices. Then $SPG(u,v)\simeq SPG(u',v')$ if and only if one of the following conditions hold under BFC: (1) $\lambda_v(u)= \lambda_{v'}(u')$ and $\lambda_u(v)= \lambda_{u'}(v')$; (2) $\lambda_v(u)= \lambda_{u'}(v')$ and $\lambda_u(v)= \lambda_{v'}(u')$.
\end{restatable}

\begin{definition}
Given a vertex $v\in V_G$ and a fixed $\delta \geq 1$, an \emph{ego shortest-path graph} (ESPG) $S_v = (V_{S_v}, E_{S_v})$ is a subgraph of $G$, where $V_{S_v} = N_{\delta}(v)$ and $E_S$ is the set of all edges that form all shortest paths between $v$ and $u\in N_{\delta}(v)$.
\end{definition}



\begin{figure}[ht]
\centering
\includegraphics[clip,width=0.65\columnwidth]{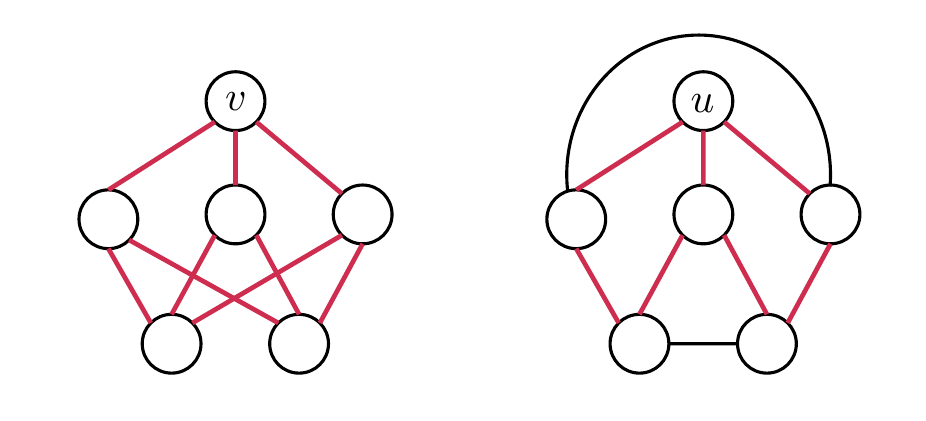}
\caption{A pair of non-isomorphic three-regular graphs. Pink edges form ESPGs ($\delta = 2$) for vertices $v$ and $u$.} 
\label{fig:esgp_examples}
\end{figure}

\begin{restatable}[]{lemma}{lvcbfsesgp}
\label{lemma:lvcbfs_esgp}
Let $v,u \in V$ be any two vertices and $\lambda(v)$ and $\lambda(u)$ be the corresponding stable colours of $v$ and $u$ by running BFC.  We have $S_v\simeq S_u$ if and only if $\lambda(v) = \lambda(u)$, where $S_v$ and $S_u$ are the ESPGs of $v$ and $u$, respectively. 

\end{restatable}

\begin{restatable}[]{lemma}{mpnnesgp}
\label{lemma:mpnn_esgp}
MPNN cannot distinguish one or more pairs of graphs that have non-isomorphic ESPGs.
\end{restatable}
\cref{fig:esgp_examples} shows a pair of graphs with non-isomorphic ESPGs that BFC is able to distinguish but MPNN cannot.





\subsection{DFS-guided Colouring}
DFS is also a fundamental graph search method for graph problems, including detecting cycles, topological sorting, and biconnectivity.
Unlike BFS whose search range grows incrementally by hop, DFS explores vertices as far as possible along each branch before backtracking. 
Before introducing a DFS-guided vertex colouring, recalling that we must have $v_j\prec v_i$ (or $i>j$) for any DFS back edge $(v_i, v_j)$, we introduce two concepts used in defining $\eta$ for DFS. 

\begin{figure}[t!]
\centering
\begin{subfigure}{.42\columnwidth}
\includegraphics[clip,width=\textwidth]{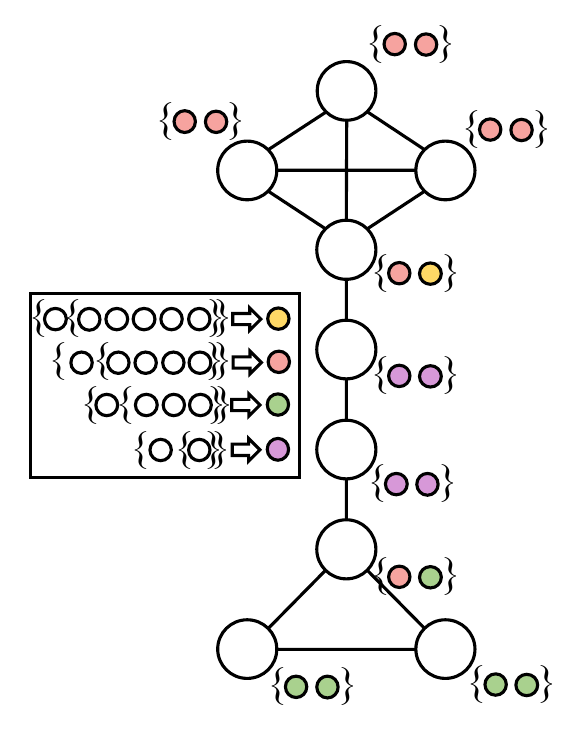}
\subcaption{}
\label{subfig:dfc1}
\end{subfigure}%
\hfill
\begin{subfigure}{.245\columnwidth}
\includegraphics[clip,width=\textwidth]{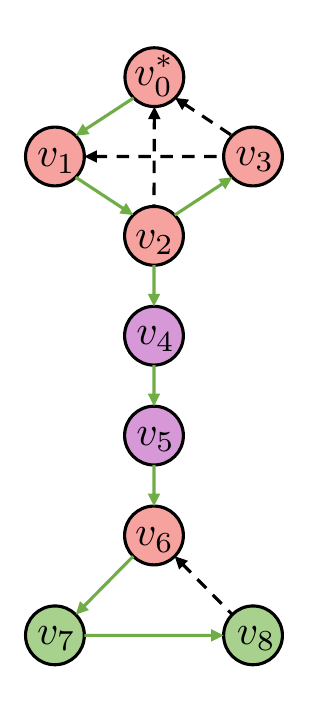}
\subcaption{}
\label{subfig:dfc2}
\end{subfigure}
\hfill
\begin{subfigure}{.245\columnwidth}
\includegraphics[clip,width=\textwidth]{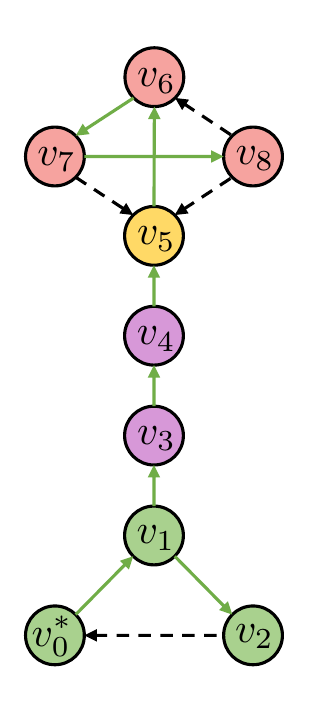}
\subcaption{}
\label{subfig:dfc3}
\end{subfigure}
\caption{
An uneven barbell graph and vertex colours after the first DFC iteration. 
\ref{subfig:dfc1} shows an uncoloured graph, where the vertex colours obtained from \ref{subfig:dfc2} and \ref{subfig:dfc3} are shown next to each vertex.
\ref{subfig:dfc2} and \ref{subfig:dfc3} show DFC with two different roots (marked with *), where each vertex is assigned a new colour by DFC. The colour map is shown on the left.
The subscripted label $v_0$, \dots, $v_5$ denote the visit sequence of each vertex, e.g. $v_1$ is visited after $v_0$.
For brevity, DFC is shown with only two roots.
} 
\label{fig:dfc}
\end{figure}

\begin{definition}[Back edge crossover]
Let $T_v$ be a search tree rooted at vertex $v\in V$ and $E_{\text{back}}^{T_v}$ be the set of back edges of $T_v$. A relation \emph{crossover} $\nmid$ on $E_{\text{back}}^{T_v}$ is defined as $\vec{e}_1 \nmid \vec{e}_2$, where $\vec{e}_1 = (v_{i_1}, v_{j_1})$ and $\vec{e}_2=(v_{i_2}, v_{j_2})$,
if one of the following four conditions is satisfied:
\begin{enumerate}
    \item[(1)]  $j_2 < j_1 < i_2 < i_1$
    \item[(2)]  $j_1 < j_2 < i_1 < i_2$
    \item[(3)]  $j_1 = j_2\text{, }i_1 \neq i_2$
    \item[(4)]  $i_1 = i_2\text{, }j_1 \neq j_2$
\end{enumerate}
\end{definition}
It is easy to see that $\nmid$ is symmetric. That is, $\vec{e}_2 \nmid \vec{e}_1$  if and only if $\vec{e}_1 \nmid \vec{e}_2$. For instance, in \cref{subfig:dfc2} we have $(v_2,v_0)\nmid (v_3,v_1)$. In \cref{subfig:edge_type_c}, we have $(v_3,v_1)\nmid (v_5,v_2)$ and $(v_5,v_0)\nmid (v_5,v_2)$. 


\begin{definition}[Back edge cover]
Let $v\in V$ be a vertex, $T_v$ be a search tree rooted at vertex $v$, $\vec{e} \in E_{\text{back}}^{T_v}$ be a back edge of $T_v$, and $\vec{e} = (v_{i}, v_{j})$. We say that $v$ is \emph{covered by} $\vec{e}$, denoted as $v \dashv \vec{e}$,
if and only if  $v \in \mathcal{P}^{T_v}_{v_jv_i}$, where $\mathcal{P}^{T_v}_{v_jv_i}$ is a path from $v_j$ to $v_i$, formed by tree edges of $T_v$.
\end{definition}
In \cref{subfig:edge_type_c}, we have $v_2 \dashv(v_3,v_1)$. In \cref{subfig:dfc3}, we have $v_1 \dashv (v_2, v_0)$.

\paragraph{Depth-first colouring (DFC).} Given a vertex $u\in V$, we define the set of back edges that cover vertex $u$ as
\begin{equation}
\label{eqn:dfc_qu}
    Q_u^{T_v} = \{ \vec{e}:{u \dashv \vec{e}, \vec{e} \in E_{\text{back}}^{T_v}} \}
\end{equation}
We then find all back edges that are transitively crossovered by $Q_u^{T_v}$. We define $D_u^{T_v,0} = Q_u^{T_v}$ for the first iteration. At the $k$-th iteration, we define
\begin{equation*}
    \Delta D_u^{T_v,k} = \{ \vec{e}_2: \vec{e}_1 \nmid \vec{e}_2, \vec{e}_1 \in D_u^{T_v, k-1}, \vec{e}_2 \in E_{\text{back}}^{T_v}  \}
\end{equation*}
\begin{equation*}
    D_u^{T_v,k} := \Delta D_u^{T_v,k}\cup D_u^{T_v, k-1}
\end{equation*}
We use $D_u^{T_v}$ to denote the set of all back edges that are transitively crossovered by edges in $Q_u^{T_v}$, i.e., when the fixed point is reached. The set of vertices covered by at least one back edge in $D_u^{T_v}$ is defined as 
\begin{equation}
\label{eqn:dfs_bu}
    B_u^{T_v} = \{o: o \dashv \vec{e}, \vec{e} \in D^{T_v}_u, o \in V\}.
\end{equation}

We design $\eta$ for DFS, denoted $\eta_{\text{dfc}}$, as
\begin{align}
\label{eqn:sigma_df}
\begin{split}
    \eta_{\text{dfc}}(u, E_{\text{tree}}^{T_v}, &E_{\text{back}}^{T_v}) = \\
    & \{o: (o,u) \in E_{\text{tree}}^{T_v}  \}
    \cup 
    \{o: o \in B_u^{T_v}\}
\end{split}
\end{align}

The part $\{o: (o,u) \in E_{\text{tree}}^{T_v}\}$ preserves vertices that lead to $u$ via tree edges. 
$\{o: o \in B_u^{T_v}\}$ preserves vertices that are covered by the same back edges covering $u$, or by back edges that transitively cover $u$.

The vertex colouring scheme defined using Equations \ref{eqn:lvc-all}, \ref{eqn:lvc}, and \ref{eqn:sigma_df} is called \textit{depth-first colouring} (DFC).
DFC is referred to as DFC-$\delta$ if the search range is limited to a $\delta$-hop neighbourhood for each $u\in V$. Because $\eta_{\text{dfc}}$ is permutation invariant, it is easy to see that DFC-$\delta$ is also permutation invariant.
\cref{fig:dfc} shows an example of colouring an uneven barbell graph using DFC. It can be seen that the vertices at two ends of the barbell share the same colour, while the vertices on the connecting path have different colours.
\begin{restatable}[]{lemma}{dfcinvariance}
\label{lemma:dfcinvariance}
$\eta_{\text{dfc}}$ is invariant under search order permutation.
\end{restatable}

\paragraph{Distinguishing biconnectivity.}
We hereby show that DFC is expressive for distinguishing graphs that exhibit the biconnectivity property. 
Graph biconnectivity is a well-studied topic in graph theory and often discussed in the context of network flow and planar graph isomorphism~\citep{Hopcroft1973-lu}. \citet{anonymous2023rethinking} first draw attention to biconnectivity in the context of GNN and show that most GNNs cannot learn biconnectivity. 

A vertex $v\in V$ is said to be a \emph{cut vertex} (or \emph{articulation point}) in $G$ if removing the vertex disconnects $G$. Thus, the removal of a cut vertex increases the number of connected components in a graph (a connected component is an inducted subgraph of $G$ in which each pair of vertices is connected via a path). Similarly, an edge $(v,u)\in E$ is a \emph{cut edge} (or \emph{bridge}) if removing $(v,u)$ increases the number of connected components. A graph is \emph{vertex-biconnected} if it is connected and does not have any cut vertices. Similarly, A graph is \emph{edge-biconnected} if it is connected and does not have any cut edges.


\begin{restatable}[]{lemma}{dfcbiconnectivity}
\label{lemma:dfcbiconnectivity}
Let $G$ and $H$ be two graphs, and 
$\{\!\!\{ \lambda(u):u\in V_G\}\!\!\}$ and $\{\!\!\{ \lambda(u'):u'\in V_H\}\!\!\}$ be the corresponding multisets of stable vertex colours of $G$ and $H$ by running DFC. Then the following statements hold:

\begin{itemize}
    \item For any two vertices $u\in V_G$ and $u'\in V_H$, if $\lambda(u) = \lambda(u')$, then $u$ is a cut vertex if and only if $u'$ is a cut vertex.
    \item For any two edges $(u_1,u_2)\in E_G$ and $(u'_1,u'_2)\in E_H$, if $\{\!\!\{ \lambda(u_1), \lambda(u_2)\}\!\!\} = \{\!\!\{ \lambda(u'_1), \lambda(u'_2)\}\!\!\}$, then $(u_1,u_2)$ is a cut edge if and only if $(u'_1,u'_2)$ is a cut edge.
    \item If $G$ is vertex/edge-biconnected but $H$ is not, then $\{\!\!\{\lambda(u): u\in V_G\}\!\!\} \neq \{\!\!\{\lambda(u'):u'\in V_H\}\!\!\}$.
\end{itemize}
\end{restatable}


\begin{restatable}[]{corollary}{vertexinbiconnectedcomponent}
\label{corollary:vertexinbiconnectedcomponent}
Let $u\in V_G$ and $u'\in V_H$ be two vertices, and $\lambda(u) = \lambda(u')$. Then $u$ is in a cycle if and only if $u'$ is in a cycle. 

\end{restatable}



\subsection{Expressivity Analysis}

\paragraph{Comparison with 1-WL.}
When $\delta=1$, it is easy to see that search trees in BFC contain only direct neighbours of the root and edges between the root and its neighbours, and there are no back edges. Thus, we have the lemma below.
\begin{restatable}[]{lemma}{onewlequal}
\label{lemma:1wlequal}
BFC-1 is equivalent to 1-WL.
\end{restatable}

When $\delta=1$, search trees in DFC contain direct neighbours of the root, edges between the root and the neighbours, and edges between the neighbours. We have the following lemma.
\begin{restatable}[]{lemma}{dfconewl}
\label{lemma:dfc1wl}
DFC-1 is more expressive than 1-WL.
\end{restatable}

\begin{restatable}[]{corollary}{bfcdistinguish}
\label{lemma:bfc_distinguish}
When $\delta > 1$, BFC-$\delta$ can distinguish one or more pairs of graphs that cannot be distinguished by 1-WL.
\end{restatable}

\begin{restatable}[]{corollary}{dfcdistinguish}
\label{lemma:dfc_distinguish}
When $\delta\geq 1$, DFC-$\delta$ can distinguish one or more pairs of graphs that cannot be distinguished by 1-WL.
\end{restatable}

\begin{figure}[h]
\centering
\includegraphics[clip,width=0.75\columnwidth]{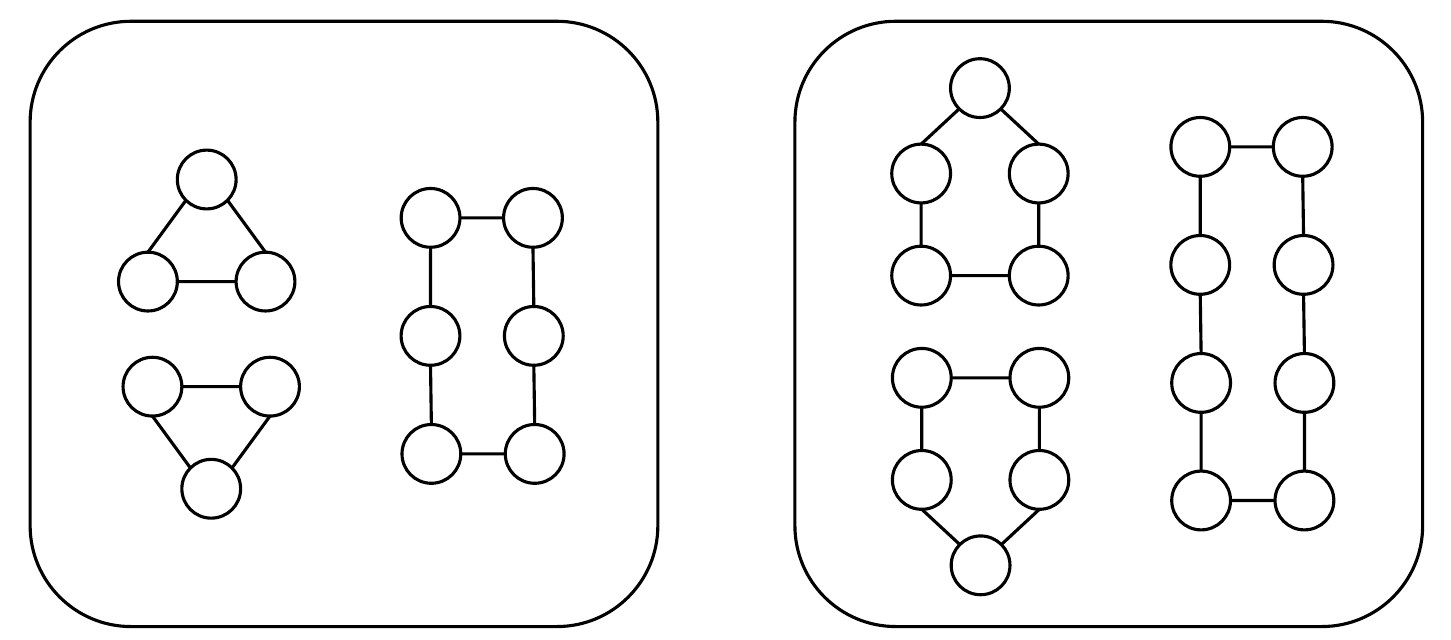}
\caption{(Left) A graph pair can be distinguished by BFC-2 and DFC-1 but not by BFC-1 and 1-WL. (Right)  A graph pair can be distinguished by BFC-3 and DFC-2 but not by BFC-2, DFC-1 and 1-WL.}
\label{fig:circle_examples}
\end{figure}

Taking a pair of graphs - one is two triangles and the other is one six-cycle - for example, these two graphs cannot be distinguished by 1-WL. However, they can be distinguished by BFC-2 and DFC-1 (\cref{fig:circle_examples}).

\paragraph{Comparison with 3-WL.}
We can also show that, regardless of the choice of $\delta$, BFC is no more expressive than 3-WL. This leads to the following theorem.

\begin{restatable}[]{thm}{bfcthreewl}
\label{thm:bfc3wl}
The expressive power of BFC-$\delta$ is strictly upper bounded by 3-WL.
\end{restatable}

\begin{figure}[h]
\centering
\includegraphics[clip,width=0.75\columnwidth]{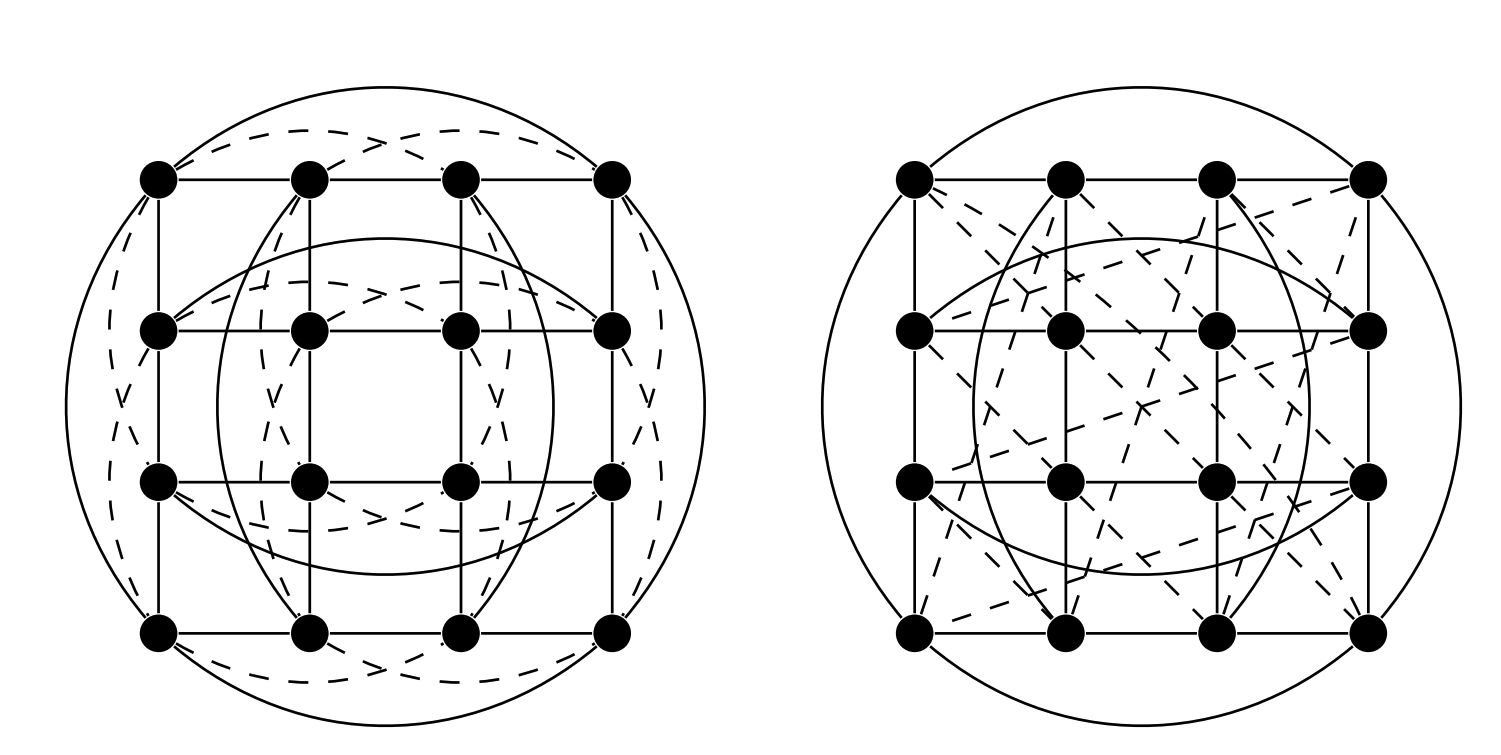}
\caption{The 4x4 Rook’s graph and the Shrikhande graph \citet{ARVIND202042}. Some edges are dashed for readability.}
\label{fig:rook_shrikhande}
\end{figure}

However, unlike BFS, DFC can distinguish graphs that cannot be distinguished by 3-WL. For example, DFC-1 can distinguish the strongly regular graph pair shown in \cref{fig:rook_shrikhande} which cannot be distinguished by 3-WL: the 4x4 Rook's graph of 16 vertices~\citep{Wagon_Weisstein} and the Shrikhande graph~\citep{Shrikhande_graph}. In the 4x4 Rook's graph, each vertex's 1-hop subgraph has a cut vertex, while there is no cut vertex in the Shrikhande graph. So according to \cref{lemma:dfcbiconnectivity}, DFC-1 can distinguish these two graphs. On the other hand, there are also graphs that 3-WL can distinguish but DFC-1 cannot. For example, the right graph pair in \cref{fig:circle_examples}. Thus, we have the following theorem.

\begin{restatable}[]{thm}{dfcthreewl}
\label{thm:dfc3wl}
The expressive powers of DFC-$\delta$ and 3-WL are incomparable.
\end{restatable}

\paragraph{Expressivity hierarchy.} The following theorem states that there exists a hierarchy among the expressive powers of BFC-$\delta$ when increasing $\delta$.

\begin{restatable}[]{thm}{bfcexpressitybeyond}
\label{thm:bfcexpressitybeyond}
BFC-$\delta$+1 is strictly more expressive than BFC-$\delta$ in distinguishing non-isomorphic graphs.
\end{restatable}

\cref{thm:bfcexpressitybeyond} implies that BFC can be used as an alternative way, separating from the WL test hierarchy, to measure the expressivity of GNNs. 

\begin{figure}[ht]
\centering
\includegraphics[clip,width=0.65\columnwidth]{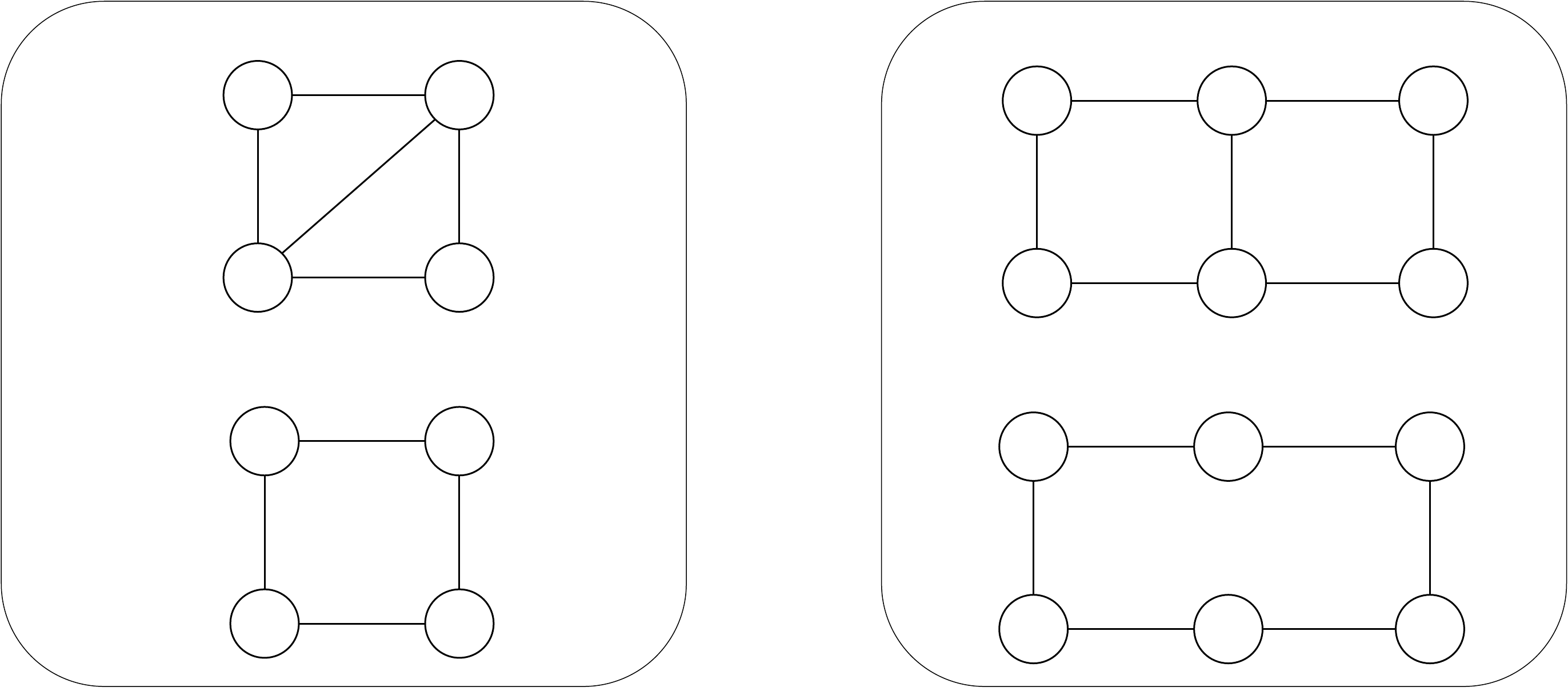}
\caption{Non-isomorphic graph pairs. A pair can be distinguished by DFC-1 but not by DFC-2 (Left). A pair can be distinguished by DFC-2 but not by DFC-3 (Right).} 
\label{fig:dfc_expressity_delta}
\end{figure}

Nevertheless, DFC-$\delta$ does not exhibit a hierarchy. \cref{fig:dfc_expressity_delta} 
depicts two pairs of non-isomorphic graph pairs: one pair can be distinguished by DFC-1 but not by DFC-2 while the other pair can be distinguished by DFC-2 but not by DFC-3. This leads to Theorem~\ref{thm:dfcexpressitynotbeyond} below. 

\begin{restatable}[]{thm}{dfcexpressitynotbeyond}
\label{thm:dfcexpressitynotbeyond}
DFC-$\delta$+1 is not necessarily more expressive than DFC-$\delta$ in distinguishing non-isomorphic graphs. 
\end{restatable}

%% file: sections/model.tex
\section{Search Guided Graph Neural Network}

We propose a graph neural network that adopts the same colouring mechanisms as BFC and DFC for feature propagation. Let $h_u^{(l)}$ be an embedding of vertex $u$ after the $l$-th layer. Our search-guided graph neural network propagates embeddings over a graph via the following update rule:
\begin{equation}
\label{eqn:gnn_layer}
  h_u^{(l+1)} = \text{MLP}\left(\left(1 + \epsilon^{(l+1)}\right)\cdot h_u^{(l)} \parallel
                        \sum_{v\in N_{\delta}(u)} h_{u\leftarrow v}^{(l+1)}
  \right)
\end{equation}
where
\begin{equation}
\label{eqn:gnn_huv}
  h_{u\leftarrow v}^{(l+1)} = \left(h_u^{(l)} +
                        \sum_{
                        w \in \eta_v(u)
                        } h_{w\leftarrow v}^{(l)}
  \right)W_{c}
\end{equation}
where, $h^{(0)}_u$ is the input vertex feature of $u$ and $\parallel$ is the concatenation operator. $h^{(l)}_{w\leftarrow v}$ is the vertex representation of $w$ obtained from a graph search rooted at $v$. When $w=v$, we have $h^{(l)}_{v\leftarrow v} = h^{(l)}_v$. Here, $W_{c}\in \mathbb{R}^{F\times H}$ is a learnable parameter matrix, where $F$ and $H$ are the dimensions of the input feature and hidden layer, respectively. Note that we use the same $W_c$ for all vertices in $\{u\in V : c=d(u,v)\}$, which have the same shortest path distance $c=d(u,v)$ to vertex $v$. For brevity, we use $\eta_v(u)$ to denote $\eta(u,E^{T_v}_{\text{tree}},E^{T_v}_{\text{back}})$.

For the model to learn an injective transformation, we use a multilayer perceptron (MLP) and a learnable scalar parameter $\epsilon$ adopted from \citet{xu2018powerful}.
The model architecture closely matches the one in Equations \ref{eqn:lvc-all} and \ref{eqn:lvc}, except that $\psi(\cdot)$ is replaced with a summation, $\phi(\cdot)$ is replaced by matrix multiplication, and MLP is used in place for $\rho(\cdot)$. 


We term the model as \textit{Search-guided Graph Neural Network}~(SGN). When $\eta_{\text{bfc}}(\cdot)$ is used in place for $\eta_v(\cdot)$, we call it SGN-BF. When $\eta_{\text{dfc}}(\cdot)$ is used, we call it SGN-DF.

\begin{restatable}[]{thm}{lvcgnn}
\label{thm:lvcgnn}
\model{} defined by the update rule in Equations \ref{eqn:gnn_layer} and \ref{eqn:gnn_huv} and with sufficiently many layers is as powerful as $LVC$.
\end{restatable}

\begin{table}[b]
    \centering
    \caption{Time and space complexity comparison.}
    \label{tab:complexity}
    \resizebox{0.9\columnwidth}{!}{
    \begin{tabular}{lcccc}
        \toprule
         & MPNN & Graphormer-GD & SGN-BF & SGN-DF \\
         \midrule
         Time & $|V|+|E|$ & $|V|^2$ & $|V|+|V|d^{\delta-1}$ & $|V|+|V|d^{2\delta}$ \\
         Space & $|V|$ & $|V|$ & $|V|d^{\delta-1}$ & $|V|d^{2\delta}$ \\
         \bottomrule
    \end{tabular}
    }
\end{table}

\paragraph{Choice of $\delta$.} As of \cref{thm:bfcexpressitybeyond}, a larger $\delta$ implies higher expressivity for \model{}-BF in distinguishing isomorphic graphs. However, increasing $\delta$ also increases the size of $\eta_v(u)$, which  makes it more expensive to compute as more aggregation operations are needed. Further, a larger $\delta$ means a larger receptive field at each layer, which is more likely to cause \textit{over-squashing}~\citep{topping22oversquashing} leading to degrade performance on vertex-level tasks, e.g. vertex classification on heterophilic graphs. 

For \model{}-DF, increasing $\delta$ does not necessarily increase expressivity (\cref{thm:dfcexpressitynotbeyond}). However, having a larger $\delta$ means that the model can detect larger biconnected components. Therefore, in practice, we combine vertex representations from $\delta=1$ with a larger $\delta$ for \model{}-DF. This guarantees that the model is more expressive than 1-WL and allows the additional $\delta$ to be fine-tuned for each dataset and task.

\paragraph{Complexity.}
$\eta_v(u)$ in \autoref{eqn:gnn_huv} only needs to be computed once, along with graph searching. Therefore, the time complexity to compute $\eta_v(u)$ is on par with the adopted graph searching. Both BFS and DFS have the worst-case complexity $O(|V|+|E|)$ for searching the whole graph. Assuming $d$ is the average vertex degree, when the search is limited to $N_{\delta}(v)$, we have the complexity $O({d}^{\delta}(1+{d}))$. We compute $\eta_v(u)$ for each $v\in V$ so in total it is $O(|V|{d}^{\delta}(1+{d}))$. Each layer in \model{} (\autoref{eqn:gnn_layer}) aggregates $|N_{\delta}(u)|$ vertex representation vectors for each vertex. Each vertex representation further aggregates $|\eta_v(u)|$ vector in \autoref{eqn:gnn_huv}, i.e., $|N_{\delta}(u)|\cdot|\eta_v(u)|$ operations, or, $O({d}^{\delta}|\eta_v(u)|)$ in total. The magnitude of $|\eta_v(u)|$ can be very different for BFS and DFS, and varies from graph to graph. In the worst case of BFS where every vertex of hop $\delta+1$ is connected to every vertex of hop $\delta$; thus $|\eta_v(u)| = \frac{1 - {d}^{\delta}}{1-{d}}$. In the worst case of DFS, $\eta_v(u)$ includes all vertices in $N_{\delta}(u)$ which has the size of ${d}^{\delta}$. We compare the time and space complexity of our model in \cref{tab:complexity} with MPNN and Graphoormer-GD~\citep{anonymous2023rethinking}.

%% file: sections/experiment.tex


\section{Experiments}
We evaluate \model{} on two prediction tasks: vertex classification and graph classification.
\model{} is implemented using Pytorch and Pytorch Geometric~\citep{fey2019fast}. Experiments are run on a single NVIDIA RTX A6000 GPU with 48GB memory.

\begin{table}[t]
\centering
\caption{Vertex classification results on homophilic datasets. }
\label{tab:vertexclassificationhomo}
    \resizebox{\columnwidth}{!}{
    \begin{tabular}{lccccc}
    \toprule
         & \amazoncomputers{} & \amazonphoto{} & \citeseer{} & \cora{} & \pubmed{} \\ \midrule
        MLP & 82.9\sd{0.4} & 84.7\sd{0.3} & 76.6\sd{0.9} & 77.0\sd{1.0} & 85.9\sd{0.2} \\
        GCN & 83.3\sd{0.3} & 88.3\sd{0.7} & 79.9\sd{0.7} & 87.1\sd{1.0} & 86.7\sd{0.3} \\
        $\text{GCN+JK}^{\dagger}$ & - & - & 74.5\sd{1.8} & 85.8\sd{0.9} & 88.4\sd{0.5} \\ 
        GAT & 83.3\sd{0.4} & 90.9\sd{0.7} & \textbf{80.5}\sd{0.7} & 88.0\sd{0.8} & 87.0\sd{0.2} \\
        APPNP & 85.3\sd{0.4} & 88.5\sd{0.3} & 80.5\sd{0.7} & 88.1\sd{0.7} & 88.1\sd{0.3} \\
        ChevNet & 87.5\sd{0.4} & 93.8\sd{0.3} & 79.1\sd{0.8} & 86.7\sd{0.8} & 88.0\sd{0.3} \\
        GPRGNN & 86.9\sd{0.3} & 93.9\sd{0.3} & 80.1\sd{0.8} & 88.6\sd{0.7} & 88.5\sd{0.3} \\ 
        BernNet & 87.6\sd{0.4} & 93.6\sd{0.4} & 80.1\sd{0.8} & 88.5 & 88.5\sd{1.0} \\ 
        $\text{H}_2\text{GCN}^{\dagger}$ & - & - & 77.1\sd{1.6} & 87.8\sd{1.4} & 89.6\sd{0.3} \\\midrule 
        \model{}-BF & 90.7 & \textbf{96.1}\sd{0.2} & 78.0\sd{1.0} & 88.7\sd{0.1} & \textbf{90.2}\sd{3.5} \\
        \model{}-DF & \textbf{90.9}\sd{0.4} & 95.2\sd{0.8}& 79.7\sd{0.7} & \textbf{89.5}\sd{0.6} & 89.5\sd{0.6}\\ \bottomrule
    \end{tabular}
    }
\vspace{1em}

\caption{Vertex classification results on heterophilic datasets.}
\label{tab:vertexclassificationhete}
    \resizebox{\columnwidth}{!}{
    \begin{tabular}{lccccc}
    \toprule
         & \wisconsin{} & \cornell{} & \texas{} & \chameleon{} & \squirrel{} \\ \midrule
        MLP & 85.3\sd{3.6} & 90.8\sd{1.6} & 91.5\sd{1.1} & 46.9\sd{1.5} & 31.0\sd{1.2} \\
        GCN & 59.8\sd{7.0}& 65.9\sd{4.4} & 77.4\sd{3.3} & 59.6\sd{2.2} & 46.8\sd{0.9} \\
        $\text{GCN+JK}^{\dagger}$ & 74.3\sd{6.4} & 74.3\sd{6.4} & 64.6\sd{8.7} & 63.4\sd{2.0}& 40.5\sd{1.6} \\ 
        GAT & 55.3\sd{8.7} & 78.2\sd{3.0} & 80.8\sd{2.1} & 63.1\sd{1.9} & 44.5\sd{0.9} \\
        APPNP & - & \textbf{91.8}\sd{2.0} & 91.0\sd{1.6} & 51.8\sd{1.8} & 34.7\sd{0.6} \\
        ChevNet & 82.6\sd{4.6} & 83.9\sd{2.1} & 86.2\sd{2.5} & 59.3\sd{1.3} & 40.6\sd{0.4} \\
        GPRGNN & - & 91.4\sd{1.8} & 93.0\sd{1.3} & 67.3\sd{1.1} & 50.2\sd{1.9} \\ 
        $\text{H}_2\text{GCN}^{\dagger}$ & 86.7\sd{4.7}& 82.2\sd{4.8} & 84.5\sd{6.8} & 59.4\sd{2.0} & 37.9\sd{2.0} \\ \midrule
        \model{}-BF & \textbf{91.2}\sd{1.0} & 89.5\sd{2.7} & 88.7\sd{4.3} & \textbf{72.8}\sd{0.2} & \textbf{59.0}\sd{0.3} \\
        \model{}-DF & 84.1\sd{3.6} &  83.2\sd{5.8} & 86.8\sd{5.2} & 56.6\sd{3.0} & 47.0\sd{1.5}\\ \bottomrule
    \end{tabular}
    }
\end{table}

\subsection{Vertex Classification}

\paragraph{Datasets.} 
We use three citation graphs, \cora{}, \citeseer{} and \pubmed{}, and two Amazon co-purchase graphs, \amazoncomputers{} and \amazonphoto{}. As shown in \citet{chien21} these five datasets are homophilic graphs on which adjacent vertices tend to share the same label. We also use two Wikipedia graphs, \chameleon{} and \squirrel{}, and two webpage graphs, \texas{} and \cornell{}, from WebKB~\citep{Pei2020}. These five datasets are heterophilic datasets on which adjacent vertices tend to have different labels.
Details about these datasets are shown in \cref{tab:statistics_vertex_classification}.

\paragraph{Setup and baselines.} We adopt the same experimental setup as \citet{bernnet21}, where each dataset is randomly split into train/validation/test set with the ratio of 60\%/20\%/20\%. In total, we use 10 random splits for each dataset, and the reported results are averaged over all splits.

We compare \model{} with seven baseline models: MLP, GCN~\citep{kipf2016semi}, GAT~\citep{velivckovic2017graph}, APPNP~\citep{KlicperaBG19}, ChebNet~\cite{defferrard2016convolutional}, GPRGNN~\citep{chien21}, and BernNet~\cite{bernnet21}.

We perform a hyperparameter search on four parameters in the following ranges:
$\text{number of layers}\in \{1, 2,3,4,5\}$,
$\text{dropout probability} \in \{0.2,0.5,0.7,0.9\}$,
$\delta\in\{1,2,3,4\}$, and
$\text{hidden layer dimension}\in\{64,128\}$.

\paragraph{Observation.}
Results on homophilic and heterophilic datasets are presented in \cref{tab:vertexclassificationhomo,tab:vertexclassificationhete}, respectively. Results marked with $\dagger$ are obtained from \citet{zhu2020beyond}, and other baseline results are taken from \citet{bernnet21}.

From \cref{tab:vertexclassificationhomo,tab:vertexclassificationhete}, we can see that \model{} generalizes to both homophilic and heterophilic graphs in vertex classification tasks. Specifically, \model{}-BF outperforms baselines in 7 out of 10 datasets, while \model{}-DF outperforms 3 out of 10. \model{}-BF performs better than \model{}-DF in heterophilic graphs. We find that the number of vertices in $N_{\delta}(v)$ grows much faster for \model{}-DF than \model{}-BF as $\delta$ increases. This can be explained as the paths to reach each $u\in N_{\delta}(v)$ from $v$ are longer in DFS than that of BFS (in BFS the path lengths are always less than or equal to $\delta$). Therefore more vertices are included in $N_{\delta}(v)$ for DFS. This further implies more vertex features are aggregated for each vertex in \model{}-DF, resulting in over-squashing which degrades the performance of \model{}-DF on heterophilic graphs.

We also list the training runtime in \cref{tab:runtime}. As expected, as $\delta$ increases, the runtime of \model{}-BF increases but stays on par with GCN. When $\delta=1$, \model{}-DF aggregates more vertices thus is slower than \model{}-BF

\begin{table}[ht]
    \centering
    \caption{Training runtime per epoch in seconds.}
    \label{tab:runtime}
    \resizebox{0.9\columnwidth}{!}{
    \begin{tabular}{lcccccc}
        \toprule
         & \shortstack{GCN\\ \phantom{} } & \shortstack{SGN-BF\\($\delta=1$)}& \shortstack{SGN-BF\\($\delta=2$)} & \shortstack{SGN-BF\\($\delta=3$)} & \shortstack{SGN-DF\\($\delta=1$)} \\
         \midrule
         \cora{} & 0.177 & 0.125 & 0.213 & 0.239 & 0.179 \\
         \pubmed{} & 0.349 & 0.224 & 0.315 & 1.271 & 0.219 \\
         \chameleon{} & 0.205 & 0.198 & 0.457 & - & 0.346 \\
         \bottomrule
    \end{tabular}
    }
\end{table}

\subsection{Graph Classification}
\paragraph{Datasets.} We evaluate \model{} on graph classification for chemical compounds, using four molecular datasets: \dd{}~\citep{Dobson2003-dd}, \proteins{}~\citep{Borgwardt2005-proteins}, \nci{}~\citep{Wale2008-nci1} and \enzymes{}~\citep{Schomburg2004-enzymes}. We also include a social dataset \imdbb{}. Following \citet{errica2019fair}, for molecular datasets, vertex features are a one-hot encoding of atom type, with the exception of \enzymes{} where additional 18 features are used, whereas for \imdbb{} the degree of each vertex is the sole vertex feature.
Details about these datasets are shown in \cref{tab:statistics_graph_classification}.

\paragraph{Setup and baselines.} 
We adopt the fair and reproducible evaluation setup from \citet{errica2019fair}, which uses an internal hold-out model selection for each of the 10-fold cross-validation stratified splits. 

We compare \model{} against five GNNs: DGCNN~\citep{zhang2018end}, DiffPool~\citep{Ying18diffpool}, ECC~\citep{Simonovsky17ECC}, GIN~\citep{xu2018powerful} and GraphSAGE~\cite{hamilton2017inductive}, as well as two variants of the contextual graph Markov model: E-CGMM~\citep{Atzeni21E-CGMM} and ICGMM~\citep{Castellana22ICGMM}. We also include a competitive structure-agnostic baseline method, dubbed \baseline{}, from \citet{errica2019fair}.

We perform the hyperparameter search:
$\text{number of layers}\in \{1, 2,3,4,5\}$,
$\text{dropout probability} \in \{0.2, 0.5, 0.7\}$,
$\delta\in\{1,2,3\}$, and
$\text{hidden layer dimension}\in\{64\}$.

\paragraph{Observation.} 
Results are presented in \cref{tab:graphclassification}. Results marked with $\ddagger$ are obtained from \citet{Castellana22ICGMM}, and other baseline results are taken from \citet{errica2019fair}.

We first observe that \model{}-DF outperforms other GNNs and CGMM variants consistently. \model{}-DF also outperforms \baseline{} on 4 out of 5 datasets. Although \model{}-BF also yields competitive results on several benchmarks, it does not perform better than \model{}-DF. This suggests that the graph properties captured by \model{}-DF, such as biconnectivity, might be useful to classify such graphs. 

\begin{table}[t]
\centering
\caption{Graph classification results.}
\label{tab:graphclassification}
    \resizebox{\columnwidth}{!}{
    \begin{tabular}{lccccc}
    \toprule
         & \dd{} & \nci{} & \proteins{} & \enzymes{} & \imdbb{}\\ \midrule
        \baseline{} & \textbf{78.4}\sd{4.5} & 69.8\sd{2.2} & 75.8\sd{3.7} & 65.2\sd{6.4} & 70.8\sd{5.0}\\
        DGCNN & 76.6\sd{4.3} & 76.4\sd{1.7} & 72.9\sd{3.5} & 38.9\sd{5.7} & 69.2\sd{3.0}\\
        DiffPool & 75.0\sd{3.5} & 76.9\sd{1.9} & 73.7\sd{3.5} & 59.5\sd{5.6} & 68.4\sd{3.3}\\
        ECC & 72.6\sd{4.1} & 76.2\sd{1.4} & 72.3\sd{3.4} & 29.5\sd{8.2} & 67.7\sd{2.8}\\
        GIN & 75.3\sd{2.9} & 80.0\sd{1.4} & 73.3\sd{4.0} & 59.6\sd{4.5} & 71.2\sd{3.9}\\
        GraphSAGE & 72.9\sd{2.0} & 76.0\sd{1.8} & 73.0\sd{4.5} & 58.2\sd{6.0} & 68.8\sd{4.5}\\
        E-CGMM\textsuperscript{$\ddagger$} & 73.9\sd{4.1} & 78.5\sd{1.7} & 73.3\sd{4.1} & - & 70.7\sd{3.8} \\
        ICGMM\textsuperscript{$\ddagger$} & 76.3\sd{5.6} & 77.6\sd{1.5} & 73.3\sd{2.9} & - & 73.0\sd{4.3}\\
        \midrule
        \model{}-BF & 76.3\sd{3.2} & 78.8\sd{2.9} & 74.0\sd{3.9} & 64.8\sd{7.2} & 71.4\sd{7.1}\\
        \model{}-DF & 78.01\sd{4.0} & \textbf{81.0}\sd{1.4} & \textbf{76.1}\sd{1.6} & \textbf{66.9}\sd{7.5} & \textbf{72.3}\sd{5.4}\\ \bottomrule
    \end{tabular}
    }
\end{table}



%% file: sections/conclusion.tex
\section{Conclusion}

Inspired by the 1-WL test, we propose a new graph colouring scheme, called \emph{local vertex colouring} (LVC). LVC iteratively refines the colours of vertices based on a graph search algorithm. LVC surpasses the expressivity limitations of 1-WL. We also prove that combining LVC with breath-first and depth-first searches can solve graph problems that cannot be solved with 1-WL test. 

Based on LVC, we propose a novel variant of graph neural network, named search-guided graph neural network (SGN). SGN is permutation invariant and
inherits the properties from LVC by adopting its colouring scheme to learn the embeddings of vertices. 
Through the experiments on a vertex classification task, we show that SGN can generalize to both homophilic and heterophilic graphs. The result of the graph classification task further verifies the efficiency of the proposed model.

%% file: sections/appendix.tex
\appendix
\input{sections/appendix/notation.tex}
\input{sections/appendix/dataset.tex}

\input{sections/appendix/proofs.tex}

%% file: sections/appendix/notation.tex
\section{Summary of Notations}
We summarise the notations used throughout the paper in \autoref{tbl:notation} below.
\begin{table}[h!]
\caption{\label{tbl:notation}Summary of notations.}
\label{sample-table}
\vskip 0.15in
 \renewcommand*{\arraystretch}{1.2}
\begin{center}
\begin{small}
\begin{tabular}{ll}
\toprule
Symbol & Description\\
\midrule
$\{\cdot\}$    & a set \\
$\{\!\!\{\cdot\}\!\!\}$    & a multiset \\
$|\cdot|$    & cardinality of a set/multiset/sequence \\
$\mathbb{P}(\cdot)$    & a power set \\
$G$, $H$ & undirected graphs\\
$V$    & a vertex set \\
$E$    & an edge set\\
$\vec{e}_{vu}$     & directed edge from vertex $v$ to $u$\\
$(v,u)$     & a vertex sequence of a directed edge\\
$d(v,u)$     & shortest distance between vertex $v$ and $u$\\
$N_{\delta}(v)$     & a set of vertices within $\delta$ hops of vertex $v$ \\
$\mathcal{P}_{vu}$     & a path from vertex $v$ to $u$ \\
$(w_0,w_1,...,w_k)$     & a vertex sequence in a path\\
$T_v$     & a search tree rooted at vertex $v$\\
$v \prec u$ &  $v$ precedes $u$ in search visiting order \\
$v \succ u$ &  $v$ succeeds $u$ in search visiting order \\
$E_{\text{back}}^{T_v}$ & a back edge set of the search tree $T_v$ rooted at vertex $v$\\
$E_{\text{tree}}^{T_v}$ & a tree edge set of the search tree $T_v$ rooted at vertex $v$\\
$G\simeq H$ &  $G$ and $H$ are isomorphic \\
$C$ &  a set of colours\\
$\lambda: V \rightarrow C$ & a colour refinement function that assigns a colour in $C$ to a vertex in $V$\\
$\lambda_v: V \rightarrow C$ & a colour refinement function that assigns a colour in $C$ to a vertex in $V$, based on a graph search rooted at vertex $v$\\
$\eta(u, E_{\text{tree}}^{T_v}, E_{\text{back}}^{T_v})$ & a function which takes a vertex $u$, a tree edge set $E_{\text{tree}}^{T_v}$ and a back edge set $E_{\text{back}}^{T_v}$ as input, outputs a vertex set \\
$\vee$ & logical or \\
$\wedge$ & logical and \\
$\parallel$ & concatenation \\
$\phi(\cdot)$, $\psi(\cdot)$, $\rho(\cdot)$ &  injective functions \\
$h_v$ &  vector representation of vertex $v$\\
$MLP$ &  multilayer perceptron\\
$\epsilon$ &  a learnable parameter \\
$W_{c}$ &  a learnable matrix with respect to the shortest path distance $c$ between two vertices \\
\bottomrule
\end{tabular}
\end{small}
\end{center}
\vskip -0.1in
\end{table}

%% file: sections/appendix/dataset.tex
\section{Dataset Statistics}

Statistics of the datasets used in our experiments are listed in \cref{tab:statistics_vertex_classification} and \cref{tab:statistics_graph_classification}.

\begin{table}[ht]
\centering
\caption{Statistics of datasets used for vertex classification.}
\label{tab:statistics_vertex_classification}
\resizebox{0.45\textwidth}{!}{%
\begin{tabular}{lcccc}
\toprule
 & \# Vertices & \# Edges & \# Features & \# Classes \\
 \midrule
\cora{} & 2708 & 5278 & 1433 & 7 \\
\citeseer{} & 3327 & 4552 & 3703 & 6 \\
\pubmed{} & 19717 & 44324 & 500 & 5 \\
\amazoncomputers{} & 13752 & 245861 & 767 & 10 \\
\amazonphoto{} & 7650 & 119081 & 745 & 8 \\
\chameleon{} & 2277 & 31371 & 2325 & 5 \\
\squirrel{} & 5201 & 198353 & 2089 & 5 \\
\texas{} & 183 & 279 & 1703 & 5 \\
\cornell{} & 183 & 277 & 1703 & 5 \\
\wisconsin{} & 251 & 466 & 1703 & 5 \\
 \bottomrule
\end{tabular}%
}
\end{table}

\begin{table}[ht]
    \centering
    \caption{Statistics of datasets used for graph classification.}
    \label{tab:statistics_graph_classification}
    \resizebox{0.5\textwidth}{!}{
    \begin{tabular}{lccccc}
        \toprule
         &\shortstack{\# Graphs }  & \shortstack{\# Classes } & \shortstack{Avg. \# \\ vertices} & \shortstack{Avg. \# \\ edges}& \shortstack{\# Features } \\
         \midrule
         \dd{} & 1178 & 2 & 284.32 & 715.66 & 89 \\
         \enzymes{} & 600 & 6 & 32.63 & 64.14 & 3 \\
         \nci{} &  4110 & 2 & 29.78 & 32.30 & 37 \\
         \proteins{} & 1113 & 2 & 39.06 & 72.82 & 3 \\
         \imdbb{}{} & 1000 & 2 & 19.77 & 96.53 & - \\
         \bottomrule
    \end{tabular}
    }
\end{table}

%% file: sections/appendix/proofs.tex
\section{Proofs}
We first introduce several lemmas that are useful for later proofs.

\begin{lemma}
\label{lemma:diff_lambda_diff_hop}
Given two pairs of vertices $(u,v)$ and $(u', v')$, if $d(u,v)\neq d(u',v')$, $d(u,v)\leq \delta$, and $d(u',v')\leq \delta$, then $\lambda_v(u) \neq \lambda_{v'}(u')$ under BFC.
\end{lemma}

\begin{proof}
We only show the case that $\lambda_v(u) \neq \lambda_{v'}(u')$ when $d(u,v) < d(u',v')$. This is because $\lambda_v(u) \neq \lambda_{v'}(u')$ can be shown for $d(u,v) > d(u',v')$ in the same way.
We prove this by contradiction and thus assume that $d(u,v)=\delta'$, $d(u',v')=\delta'+\Delta$, and $\lambda_v(u) = \lambda_{v'}(u')$, where $\delta'\leq \delta$ and $\Delta \geq 1$. 

Since $\lambda_v(u) = \lambda_{v'}(u')$, by Equations~\ref{eqn:lvc-all} and \ref{eqn:lvc}, we know that $\lambda^{l}(u) = \lambda^{l}(u')$ must hold for any $l\geq \delta'$; otherwise, the injectivitity of the functions $\rho$ and $\phi$ would lead to $\lambda_v(u) \neq \lambda_{v'}(u')$, contradicting with the assumption.

Then, by Equations~\ref{eqn:lvc-all} and \ref{eqn:lvc} again, we have the following for any $l\geq \delta'$:
\begin{align}
   \psi\left( \{\!\!\{ \lambda^{l}_{v}(w) : w \in \eta_{\text{bfc}}(u, E_{\text{tree}}^{T_v}, E_{\text{back}}^{T_v}) \}\!\!\}\right) = &  \psi\left(\{\!\!\{ \lambda^{l}_{v'}(w') : w' \in \eta_{\text{bfc}}(u', E_{\text{tree}}^{T_{v'}}, E_{\text{back}}^{T_{v'}}) \}\!\!\}\right).
\end{align}
Because $\psi$ is injective, the above leads to
\begin{align}
  |\eta_{\text{bfc}}(u, E_{\text{tree}}^{T_v}, E_{\text{back}}^{T_v})| = &  |\eta_{\text{bfc}}(u', E_{\text{tree}}^{T_{v'}}, E_{\text{back}}^{T_{v'}})| \text{  and  } d(u,u)=d(u',u')=0.
\end{align}
Again we know that $\lambda^{l-1}(w) = \lambda^{l-1}(w')$ because of the injectivity of $\rho$ and $\phi$. Similarly, we have the following 
\begin{align}
  |\eta_{\text{bfc}}(w, E_{\text{tree}}^{T_v}, E_{\text{back}}^{T_v})| = &  |\eta_{\text{bfc}}(w', E_{\text{tree}}^{T_{v'}}, E_{\text{back}}^{T_{v'}})| \text{  and  } d(u,w)=d(u',w')=1.
\end{align}

The above can be shown recursively for all vertices in $SPG(u,v)$ and $SPG(u',v')$. So we know that, for any vertex $z'$ in $SPG(u',v')$, there must exist a vertex $z$ in $SPG(u,v)$ such that the following conditions hold:
\begin{align}
  |\eta_{\text{bfc}}(z, E_{\text{tree}}^{T_v}, E_{\text{back}}^{T_v})| = &  |\eta_{\text{bfc}}(z', E_{\text{tree}}^{T_{v'}}, E_{\text{back}}^{T_{v'}})| \text{  and  } d(u,z)=d(u',z').
\end{align}
However, since $d(u,v)=\delta'$ and $d(u',v')=\delta'+\Delta$, there must exist at least one vertex $z'$ in $SPG(u',v')$ with $\delta'<d(u',z')\leq\delta'+\Delta$, which cannot satisfy the above conditions. This implies that $\lambda_v(u) = \lambda_{v'}(u')$ does not hold, contradicting our assumption. The proof is done.

\end{proof}

\begin{lemma}
\label{lemma:equal_neighbour_size}
$|N_{\delta}(v)| = |N_{\delta}(u)|$ if $\lambda(v)=\lambda(u)$. 
\end{lemma}
\begin{proof}
 According to \cref{eqn:lvc-all}, when $\lambda(v)=\lambda(u)$, we have
\begin{equation*}
    \phi\left(\lambda(v), \psi\left\{\!\!\{\lambda_{w_1}(v): w_1\in N_{\delta}(v)\}\!\!\}\right)\right) =
    \phi\left(\lambda(u), \psi\left\{\!\!\{\lambda_{w_2}(u): w_2\in N_{\delta}(u)\}\!\!\}\right)\right).
\end{equation*}
Because $\phi(\cdot)$ and $\psi(\cdot)$ are injective, we have 
\begin{equation*}
    \{\!\!\{\lambda_{w_1}(v): w_1\in N_{\delta}(v)\}\!\!\} =
    \{\!\!\{\lambda_{w_2}(u): w_2\in N_{\delta}(u)\}\!\!\}.
\end{equation*}
Hence, the number of elements should be the same for the multisets on the left and right sides. This leads to $|N_{\delta}(v)| = |N_{\delta}(u)|$. The proof is done.
\end{proof}

\bfcspg*

\begin{proof}
We first show $SPG(u,v)\simeq SPG(u',v')$ if one of the two conditions holds.

We only show the statement holds when the first condition holds, i.e. ``$SPG(u,v)\simeq SPG(u',v')$ if $\lambda_v(u) = \lambda_{v'}(u')$ and $\lambda_u(v) = \lambda_{u'}(v')$". The case for the second condition, i.e. ``$SPG(u,v)\simeq SPG(u',v')$ if $\lambda_v(u) = \lambda_{u'}(v')$ and $\lambda_u(v) = \lambda_{v'}(u')$", can be shown in the same way.

Assuming $\lambda_v(u) = \lambda_{v'}(u')$ and $\lambda_u(v) = \lambda_{u'}(v')$, we show $SPG(u,v)\simeq SPG(u',v')$  by induction.
\begin{itemize}
    \item When $d(u,v)=0$ and $d(u',v')=0$, we must have $u=v$ and $u'=v'$. Accordingly, both $SPG(u,v)$ and $SPG(u',v')$ contain only one node. Thus we must have $SPG(u,v)\simeq SPG(u',v')$.
    \item When $d(u,v)=1$ and $d(u',v')=1$, we have $(u,v)\in E$ and $(u',v')\in E$. Then both $SPG(u,v)$ and $SPG(u',v')$ contain only one edge. Thus we must have $SPG(u,v)\simeq SPG(u',v')$.
    \item When $d(u,v)=2$ and $d(u',v')=2$, $SPG(u,v)$ and $SPG(u',v')$ can only have one isomorphic type: assuming there are total $N$ vertices in $SPG(u,v)$ and $SPG(u',v')$, there are $N-2$ vertices adjacent to both $u$/$u'$ and $v$/$v'$. It is easy to see $SPG(u,v)\simeq SPG(u',v')$.
    
    \item Now assume that the statement ``$SPG(u,v)\simeq SPG(u',v')$ if $\lambda_v(u)= \lambda_{v'}(u')$ and $\lambda_u(v)= \lambda_{u'}(v')$ under BFC" holds for any two pairs of vertices $(u,v)$ and $(u',v')$ when $d(u,v)=d(u',v')\leq \Delta$. We want to show that this statement will hold for the case $d(u,v)=d(u',v')= \Delta+1$. It is easy to see that $\lambda_v(u)= \lambda_{v'}(u')$ if and only if $\lambda_u(v)= \lambda_{u'}(v')$, so we just need to show $SPG(u,v)\simeq SPG(u',v')$ if $\lambda_v(u) = \lambda_{v'}(u')$.
    
    When $d(u,v)=\Delta+1$, we may express $SPG(u,v)$ as a tree rooted at vertex $u$, which has a number of children $\{\!\!\{SPG(u_1,v),\dots, SPG(u_q,v)\}\!\!\}$ where $d(u_i,v)=\Delta$ for $1\leq i\leq q$ and $(u,u_i)\in E$. Accordingly, we may express $SPG(u',v')$ as a tree rooted at vertex $u'$, which has a number of children $\{\!\!\{SPG(u'_1,v'),\dots, SPG(u'_q,v)\}\!\!\}$ where $d(u'_i,v')=\Delta$ for $1\leq i\leq p$ and $(u',u'_i)\in E$.
    
    Because $\lambda_v(u) = \phi(\lambda(u), \psi(\{\!\!\{ \lambda_v(u_1), \dots, \lambda_v{u_q}\}\!\!\}))$ and $\lambda_{v'}(u') = \phi(\lambda(u'), \psi(\{\!\!\{ \lambda_{v'}(u'_1), \dots, \lambda_{v'}(u'_p)\}\!\!\}))$,  we must have $p=q$ and $\{\!\!\{\lambda_v(u_1),\dots, \lambda_v(u_q)\}\!\!\} = \{\!\!\{\lambda_{v'}(u'_1),\dots, \lambda_{v'}(u'_p)\}\!\!\}$.
    Without loss of generality, assuming $\lambda_v(u_1)=\lambda_v'(u'_q)$, \dots, $\lambda_v(u_p)=\lambda_v'(u'_p)$, 
    by our assumption for the case $d(u,v)=d(u',v')\leq \Delta$,
    we have $SPG(u_1,v)\simeq SPG(u'_1, v')$, \dots, $SPG(u_q,v)\simeq SPG(u'_p, v')$. This implies $\{\!\!\{SPG(u_1,v),\dots, SPG(u_q,v)\}\!\!\}\simeq \{\!\!\{SPG(u'_1,v'),\dots, SPG(u'_p,v')\}\!\!\}$. Thus, we must have $SPG(u,v)\simeq SPG(u',v')$.
    So the statement ``$SPG(u,v)\simeq SPG(u',v')$ if $\lambda_v(u)= \lambda_{v'}(u')$ and $\lambda_u(v)= \lambda_{u'}(v')$ under BFC" holds  for the case $d(u,v)=d(u',v')= \Delta+1$.
\end{itemize}

Now we show $SPG(u,v)\simeq SPG(u',v')$ only if one of the two conditions holds.
Assuming $SPG(u,v)\simeq SPG(u',v')$, we show this by induction.

\begin{itemize}
    \item When $d(u,v)=0$ and $d(u',v')=0$, we must have $u=v$ and $u'=v'$. Accordingly, both $SPG(u,v)$ and $SPG(u',v')$ contain only one node. Thus both conditions hold.
    \item When $d(u,v)=1$ and $d(u',v')=1$, we have $(u,v)\in E$ and $(u',v')\in E$. Then both $SPG(u,v)$ and $SPG(u',v')$ contain only one edge. Thus both conditions hold.
    \item When $d(u,v)=2$ and $d(u',v')=2$, $SPG(u,v)$ and $SPG(u',v')$ can only have one isomorphic type: assuming there are total $N$ vertices in $SPG(u,v)$ and $SPG(u',v')$, there are $N-2$ vertices adjacent to both $u$/$u'$ and $v$/$v'$. It is easy to see both conditions hold.
    
    \item Now assume that the statement holds for the first condition, i.e. ``$SPG(u,v)\simeq SPG(u',v')$ only if $\lambda_v(u)= \lambda_{v'}(u')$ and $\lambda_u(v)= \lambda_{u'}(v')$ under BFC" holds for any two pairs of vertices $(u,v)$ and $(u',v')$ when $d(u,v)=d(u',v')\leq \Delta$. We want to show that this statement also hold for the case $d(u,v)=d(u',v')= \Delta+1$. 
    
    It is easy to see that $\lambda_v(u)= \lambda_{v'}(u')$ if and only if $\lambda_u(v)= \lambda_{u'}(v')$, so we just need to show $SPG(u,v)\simeq SPG(u',v')$ only if $\lambda_v(u) = \lambda_{v'}(u')$. 
    Similar with before, when $d(u,v)=\Delta+1$, we express $SPG(u,v)$ as a tree rooted at vertex $u$, which has a number of children $\{\!\!\{SPG(u_1,v),\dots, SPG(u_q,v)\}\!\!\}$ where $d(u_i,v)=\Delta$ for $1\leq i\leq q$ and $(u,u_i)\in E$. Accordingly, we express $SPG(u',v')$ as a tree rooted at vertex $u'$, which has a number of children $\{\!\!\{SPG(u'_1,v'),\dots, SPG(u'_q,v)\}\!\!\}$ where $d(u'_i,v)=\Delta$ for $1\leq i\leq p$ and $(u',u'_i)\in E$.    
    Because $SPG(u,v)\simeq SPG(u',v')$, we must have $p=q$. This further implies $\{\!\!\{SPG(u_1,v),\dots, SPG(u_q,v)\}\!\!\}\simeq \{\!\!\{SPG(u'_1,v'),\dots, SPG(u'_p,v')\}\!\!\}$. By our assumption for the case $d(u,v)=d(u',v')\leq \Delta$, we have $\{\!\!\{\lambda_v(u_1),\dots, \lambda_v(u_q)\}\!\!\}= \{\!\!\{\lambda_{v'}(u'_1),\dots, \lambda_{v'}(u'_p)\}\!\!\}$, which further leads to $\lambda_v(u) = \lambda_{v'}(u')$. So the statement ``$SPG(u,v)\simeq SPG(u',v')$ only if $\lambda_v(u)= \lambda_{v'}(u')$ and $\lambda_u(v)= \lambda_{u'}(v')$ under BFC" holds  for the case $d(u,v)=d(u',v')= \Delta+1$.

    Now assume that the statement holds for the second condition, i.e. ``$SPG(u,v)\simeq SPG(u',v')$ only if $\lambda_v(u)= \lambda_{u'}(v')$ and $\lambda_u(v)= \lambda_{v'}(u')$ under BFC" holds for any two pairs of vertices $(u,v)$ and $(u',v')$ when $d(u,v)=d(u',v')\leq \Delta$. We can show the statement also holds  for the case $d(u,v)=d(u',v')= \Delta+1$ in the same way as before.
\end{itemize}
The proof is done.
\end{proof}

\lvcbfsesgp*

\begin{proof}
We first show that, if $\lambda(v)=\lambda(u)$, then we have $S_v\simeq S_u$.
We prove this by induction:
\begin{itemize}
    \item When $\delta=0$, there is only one vertex in $S_v$ and $S_u$, it is trivial to see $S_v\simeq S_u$.
    \item When $\delta=1$, there are only $v$ and its direct neighbours in $S_v$, and only $u$ and its direct neighbours $S_u$. $v$ and $u$ must have the same degree for $\lambda(v)=\lambda(u)$ to hold, so we have $S_v\simeq S_u$.
    \item Now assume that the statement ``$S_v\simeq S_u$ if $\lambda(v)= \lambda(u)$ under BFC" holds for any two vertices $v$ and $u$ when $\delta\leq \Delta$. We want to show that this statement will hold for the case $\delta= \Delta+1$.     When $\delta=\Delta+1$, we only consider the case where $S_v\simeq S_u$ for $\delta=\Delta$; otherwise, we immediately have $\lambda(v) \neq \lambda(u)$ according to \cref{lemma:diff_lambda_diff_hop}. Assuming that there are $q$ vertices $\{v_1, \dots, v_q\}$ in $N_{\Delta+1}(v)\setminus N_{\Delta}(v)$ and $p$ vertices $\{u_1, \dots, u_p\}$ in $N_{\Delta+1}(u)\setminus N_{\Delta}(u)$, where $p\geq 1$ and $q\geq 1$. If $p \neq q$ we have $\lambda(v) \neq \lambda(u)$. So we only consider the case where $p=q$.
    In this case, for $\lambda(v)= \lambda(u)$ to hold, we must have $\{\!\!\{\lambda_{v_1}(v),\dots,\lambda_{v_1}(v)\}\!\!\} = \{\!\!\{\lambda_{u_1}(u),\dots, \lambda_{u_p}(u)\}\!\!\}$. Then we have $\{\!\!\{SPG(v_1,v),\dots, SPG(v_q,v)\}\!\!\} \simeq \{\!\!\{SPG(u_1,u),\dots, SPG(u_p,u')\}\!\!\}$ according to \cref{lemma:bfc2spg}. Thus, we must have $S_v\simeq S_u$.

\end{itemize}

We then show that, if $S_v\simeq S_u$ we have $\lambda(v)=\lambda(u)$.
We prove this by induction:
\begin{itemize}
    \item When $\delta=0$, there is only one vertex in $S_v$ and $S_u$, it is trivial to see $\lambda(v)=\lambda(u)$.
    \item When $\delta=1$, there are only $v$ and its direct neighbours in $S_v$, and only $u$ and its direct neighbours $S_u$. $v$ and $u$ must have the same degree for $S_v\simeq S_u$ to hold, so we have $\lambda(v)=\lambda(u)$.
    \item Now assume that the statement ``$\lambda(v)= \lambda(u)$ if $S_v\simeq S_u$ under BFC" holds for any two vertices $v$ and $u$ when $\delta\leq \Delta$. We want to show that this statement will hold for the case $\delta= \Delta+1$. 
    Assuming that there are $q$ vertices $\{v_1, \dots, v_q\}$ in $N_{\Delta+1}(v)\setminus N_{\Delta}(v)$ and $p$ vertices $\{u_1, \dots, u_p\}$ in $N_{\Delta+1}(u)\setminus N_{\Delta}(u)$, where $p\geq 1$ and $q\geq 1$. If $p \neq q$ we have $S_v\not\simeq S_u$. 
    So we only consider the case where $p=q$. 
    Because $S_v\simeq S_u$ for $\delta=\Delta+1$,
    we know $\{\!\!\{SPG(v_1,v),\dots, SPG(v_q,v)\}\!\!\} \simeq \{\!\!\{SPG(u_1,u),\dots, SPG(u_p,u')\}\!\!\}$. According to \cref{lemma:bfc2spg}, we must have $\{\!\!\{\lambda_{v_1}(v),\dots, \lambda_{v_q}(v)\}\!\!\} = \{\!\!\{\lambda_{u_1}(u),\dots, \lambda_{u_p}(u)\}\!\!\}$. Thus, we have $\lambda(v)= \lambda(u)$.
\end{itemize}
The proof is done.
\end{proof}


\mpnnesgp*
\begin{proof}
Consider vertices $v$ and $u$ and their 1-hop neighbour vertex sets $N_1(v)$ and $N_1(u)$, respectively. We use $\lambda_{\text{WL}}(\cdot)$ to denote the colour mapping of 1-WL.
We show an example where $\lambda_{\text{WL}}(v) = \lambda_{\text{WL}}(u)$ but $S_v\not\simeq S_u$. In \cref{fig:esgp_examples}, $S_v$ and $S_u$ are non-isomorphic; however, we can see $\lambda_{\text{WL}}(v)=\lambda_{\text{WL}}(u)$ because each vertex is adjacent to the same number of vertices. We know from \citet{xu2018powerful} that MPNN's expressivity is upper-bounded by 1-WL; thus, the proof is done.
\end{proof}

\dfcinvariance*

\begin{proof}
By showing that $\eta_{\text{dfc}}$ is invariant to vertex search orders, we can prove that $\eta_{\text{dfc}}$ is permutation invariant.

Note that for DFS rooted at vertex $v$, there are multiple search trees if and only if there exist back edges. For a DFS tree $T_v$, any alternative DFS tree can be formed by changing a tree edge to a back edge and swap their directions. If there are no back edges in a DFS tree, the DFS tree is canonical such that the output of $\eta_{\text{dfc}}$ does not change. So below we only consider the case where back edges exist. 

For a non-root vertex $u$ in $G$, there must be one and only one tree edge leading to $u$, i.e. exact one vertex in $\{o:(o,u)\in E_{\text{tree}}^{T_v} \}$. If $Q_u^{T_v}$ in \cref{eqn:dfc_qu} is empty, $u$ is not covered by a back edge and $u$ is not in any cycle because a cycle is formed by at least one back edge~\citep{Cormen_algointro}. Since $u$ is not in any cycle, the vertex $w$ precedes $u$ in $T_v$ is deterministic.

Now we consider the case where $Q_u^{T_v}$ is not empty, which has two further cases. Let $w$ be a vertex preceding $u$ in the tree edge $(w,u)$. 

\begin{itemize}
    \item The first case is when $w$ and $u$ are covered together by a back edge, or by two back edges that crossover each other. In this case, $u$ and $w$ are in the same cycle(s), so $w$ appears in $B_u^{T_v}$. For any alternative search tree to $T_v$, $w$ and $u$ must also be covered together by a back edge, or by two back edges that crossover each other. So the union $\{o: (o,u) \in E_{\text{tree}}^{T_v}   \}
\cup  \{o: o \in B_u^{T_v}\}$ in \cref{eqn:sigma_df} remains the same for all possible search trees. 

\item The second case is when $w$ and $u$ are not covered by a back edge, or are covered by two back edges that do not crossover each other. In this case, $(w,u)$ appears as a tree edge in all possible search trees rooted at $v$, because the tree edge $(w,u)$ is not in any cycles. In this case, $D^{T_v}_u$ remains the same for all search trees, which makes the union $\{o: (o,u) \in E_{\text{tree}}^{T_v}  \}
\cup  \{o: o \in B_u^{T_v}\}$ invariant to traversal order. 
\end{itemize}
Thus, $\eta_{\text{dfc}}$ is invariant to vertex traversal order. The proof is done.
\end{proof}

In the following, we introduce a lemma that is useful for proving \cref{lemma:dfcbiconnectivity}.

\begin{lemma}
\label{lemma:binnect_back_edge_cover}
Each vertex in a biconnected component is covered by at least one DFS back edge.
\end{lemma}
\begin{proof}
A biconnected component is an induced subgraph of $G$ which stays connected by removing any one vertex. Therefore, each vertex in a biconnected component should participate in at least one cycle.
We know a vertex forms a cycle if and only if it is covered by a DFS back edge. Hence, vertices in a biconnected component is covered by at least one DFS back edge.
\end{proof}

\dfcbiconnectivity*
\begin{proof}
We prove the statements in \cref{lemma:dfcbiconnectivity} one by one.
\begin{itemize}
    \item By the definition that a vertex $u$ is a cut vertex of $G$ if removing $u$ increases the number of connected components, \citet{Hopcroft1973-lu} shows that $u$ is a cut vertex if one of the following two conditions is true:
        \begin{enumerate}
            \item $u$ is not the root of a DFS tree, and it has a child $c$ such that no vertex in the subtree rooted with $c$ has a back edge to one of the ancestors in the DFS tree of $u$.
            \item $u$ is the root of a DFS tree, and it has at least two children.
        \end{enumerate}
    \cref{eqn:lvc-all} only computes vertex colour based on search trees where $u$ is not the root, so we can just focus on condition 1. There are two types of cut vertex: one that does not form biconnected components, called the \emph{type-1 cut vertex} (e.g. $v_4$ and $v_5$ in \cref{subfig:dfc2}), and the other that forms biconnected components, called the \emph{type-2 cut vertex} ($v_2$ and $v_6$ in \cref{subfig:dfc2}). We first show that the first statement holds for the type-1 cut vertex.
    
    For a type-1 cut vertex $u$, it is easy to see that we have $D^{T_v}_u = \varnothing$ because of \cref{lemma:binnect_back_edge_cover}. Hence, $\eta_{\text{dfc}}(u,E_{\text{tree}}^{T_v},E_{\text{back}}^{T_v})$ returns a single vertex set containing $w$, where $w$ is the vertex preceding $u$ in the tree edge $(w,u)$. Note that both a type-1 cut vertex and a leaf vertex (a vertex without children) obtain a single vertex set from \cref{eqn:sigma_df}; however, according to Condition 1, a leaf vertex is not a vertex tree. We show that if $u$ is a cut vertex and $x$ is a leaf vertex, $\lambda_G(u)\neq \lambda_G(x)$. We show this by contradiction. A leaf vertex has a degree of 1 while a cut vertex must have a degree larger than 1 (because it connects at least two biconnected components). For $\lambda_G(u) = \lambda_G(x)$ to hold, according to \cref{eqn:lvc-all}, there must be the same number of vertices in $N_{\delta}(u)$ and $N_{\delta}(x)$. 
    Because $x$ is a leaf vertex, there is only one vertex, $w_1$, that contributes to the colour of $u$ without involving other vertices.
    Because $u$ is a cut vertex, there are at least two vertices, $w'_1$ and $w'_2$, that are adjacent to $u$.
    Assuming $\lambda_{w_1}(x) = \lambda_{w'_1}(u)$, there must exist another vertex $ w_2 \in N_{\delta}(x)$ such that $\lambda_{w_2}(x) = \lambda_{w'_2}(u)$ and $w_2$ does not have an edge with $x$. For simplicity, we omit the superscript in \cref{eqn:lvc}, and unpack $\lambda_{w_2}(x)$ and $\lambda_{w'_2}(u)$. We then have 
    $\lambda_{w_2}(x) = \phi \big(  \lambda(x), \psi\left( \{\!\!\{\lambda_{w_2}(w_1)\}\!\!\} \right) \big)$
    and $\lambda_{w'_2}(u) = \phi\big( \lambda(u), \psi\left( \{\!\!\{\lambda_{w'_2}(w'_2)\}\!\!\} \right) \big)$.
    Hence, we need $\lambda_{w_2}(w_1) = \lambda_{w'_2}(w'_2)$.
    Because $\lambda_{w_2}(w_1) = \phi\big(\lambda(w_1),  \psi(\{\!\!\{ \lambda_{w_2}(w_2) \}\!\!\})\big)$,  $\lambda_{w_2}(w_2) = \lambda(w_2)$, and $\lambda_{w_2}(w'_2) = \lambda(w'_2)$, 
    we need to have $\lambda(w'_2) = \phi\big(\lambda(w_1), \psi(\{\!\!\{ \lambda_{w_2}(w_2) \}\!\!\})\big)$ which obviously does not hold. Therefore, if $u$ is a cut vertex and $x$ is a leaf vertex, $\lambda_G(u)\neq \lambda_G(x)$.

    For a type-2 cut vertex $u$, it is obvious that $D^{T_v}_u \neq \varnothing$ and thus the colour of $u$ is clearly different from any leaf vertices or any type-1 cut vertices. So now we just need to show $\lambda_G(u) \neq \lambda_H(x)$ when $x$ is a non-cut vertex in a biconnected component. 
    Because $u$ is a type-2 cut vertex, $\eta_{\text{dfc}}(u,E_{\text{tree}}^{T_v},E_{\text{back}}^{T_v})$ includes all vertices in the biconnected components in which $u$ participates (at least two biconnected components), while for a non-cut vertex $x$, $\eta_{\text{dfc}}(u,E_{\text{tree}}^{T_v},E_{\text{back}}^{T_v})$ includes only vertices in a single biconnected component. 
    So a type-2 cut vertex has a different number of vertices in $\eta_{\text{dfc}}(u,E_{\text{tree}}^{T_v},E_{\text{back}}^{T_v})$ comparing with a non-cut vertex.
    Therefore, it is easy to see that $\lambda_G(u) \neq \lambda_H(x)$ when $u$ is a type-2 cut vertex and $x$ is a non-cut vertex in a biconnected component.

    Hence, if $\lambda_G(u)=\lambda_G(x)$, and $x$ is a cut vertex if and only if $u$ is a vertex. The proof for the first statement is done.

    \item According to Tarjan's algorithm for finding cut edges~\citep{Tarjan1974-eb}, in a DFS tree, a tree edge $(u_1, u_2)$ is a cut edge if there is a path from $u_1$ to $u_2$ and every path from $u_1$ to $u_2$ contains edge $(u_1, u_2)$. This can be interpreted as follows: $(u_1, u_2)$ is a cut edge if $u_1$ and $u_2$ are not covered by the same back edge; otherwise, these back edges do not crossover each other. We prove the second statement by contradiction. 
    If $(u_1, u_2)$ is a cut edge and 
    $(x_1, x_2)$ is not a cut edge, assuming 
    $\{\!\!\{ \lambda(u_1), \lambda(u_2)\}\!\!\} = \{\!\!\{ \lambda(x_1), \lambda(x_2)\}\!\!\}$, then $x_1$ and $x_2$ must be covered either by the same back edge or by different back edges that do not crossover each other. Hence, based on \cref{eqn:sigma_df}, $\eta_{\text{dfc}}(x_1,E_{\text{tree}}^{T_v},E_{\text{back}}^{T_v}) = \eta_{\text{dfc}}(x_2,E_{\text{tree}}^{T_v},E_{\text{back}}^{T_v})$. Since $(u_1, u_2)$ is a cut edge, we know $\eta_{\text{dfc}}(u_1,E_{\text{tree}}^{T_v},E_{\text{back}}^{T_v}) \neq \eta_{\text{dfc}}(u_2,E_{\text{tree}}^{T_v},E_{\text{back}}^{T_v})$. This contradicts to $\{\!\!\{ \lambda(u_1), \lambda(u_2)\}\!\!\} = \{\!\!\{ \lambda(x_1), \lambda(x_2)\}\!\!\}$; thus $(x_1, x_2)$ must be a cut edge. We can swap $(x_1, x_2)$ and $(u_1, u_2)$ in the proof to show if $(x_1, x_2)$ is a cut edge, $(u_1, u_2)$ must also be a cut edge. The proof is done for the second statement.

    \item For a graph $G$ to be cut vertex/edge connected, in a DFS tree, there must be a set of back edges that crossover each other and cover all vertices of $G$. Therefore, if $\eta_{\text{dfc}}(u,E_{\text{tree}}^{T_v},E_{\text{back}}^{T_v})$ includes all vertices of $G$, then we have $\eta(u_0,E_{\succ}, E_{\prec}) = \eta(u_1,E_{\succ}, E_{\prec}) = \dots = \eta(u_{N-1},E_{\succ}, E_{\prec})$ for $u_0,u_1,\dots,u_{N-1} \in V_G$. If $H$ is not cut vertex/edge connected, then there must exist at least one vertex $w\in V_H$ such that $w$ is not covered by the same set of crossover back edges that cover other vertices. Thus, 
    $\eta_{\text{dfc}}(w,E_{\text{tree}}^{T_v},E_{\text{back}}^{T_v}) \neq \eta_{\text{dfc}}(x,E_{\text{tree}}^{T_v},E_{\text{back}}^{T_v})$ for $x \in V_H\setminus \{w\}$. We have $\{\!\!\{\lambda(u): u\in V_G\}\!\!\} \neq \{\!\!\{\lambda(u'):u'\in V_H\}\!\!\}$ if $G$ is vertex/edge-biconnected but $H$ is not. The proof is done for the third statement.
    
\end{itemize}
\end{proof}




\vertexinbiconnectedcomponent*

\begin{proof}

    From the proofs of \cref{lemma:dfcbiconnectivity,lemma:binnect_back_edge_cover}, we know that, if $u$ is in a cycle, $u$ is covered by at least one back edge. So $\eta_{\text{dfc}}(u,E_{\text{tree}}^{T_v},E_{\text{back}}^{T_v})$ contains more than 2 vertices. Since $\lambda(u)=\lambda(u')$, $\eta_{\text{dfc}}(u',E_{\text{tree}}^{T_v},E_{\text{back}}^{T_v})$ also needs to contain more than two vertices, which indicates $u'$ is in a cycle.
    
\end{proof}

\onewlequal*

\begin{proof}
When $\Delta=1$, the vertex colouring function in \cref{eqn:lvc-all} becomes
\begin{equation}
    \lambda^{i+1}(u):= \rho(\lambda^i(u), \{\!\!\{\lambda^{i+1}_v(u)| (u,v)\in E\}\!\!\}).
\end{equation}
Accordingly, we have \cref{eqn:lvc}
\begin{equation}
   \lambda^{i+1}_v(u):= \phi\left(\lambda^i(u), \psi(\{\!\!\{\lambda^{i}(v)\}\!\!\})\right),
\end{equation}
which leads to
\begin{equation}
\label{eqn:lemma_4_6}
    \lambda^{i+1}(u) := \rho\left(\lambda^i(u), \{\!\!\{\phi\left(\lambda^i(u), \psi\left(\{\!\!\{\lambda^{i}(v)\}\!\!\}\right)\right)|(u,v) \in E\}\!\!\}\right).
\end{equation}
Elements in the multiset $\{\!\!\{\phi\left(\lambda^i(u), \psi\left(\{\!\!\{\lambda^{i}(v)\}\!\!\}\right)\right)|(u,v) \in E\}\!\!\}$ only differ in the input of $\psi(\cdot)$. Thus, we can simplify it by defining a new injective function $\kappa(\lambda^i(v)) = \phi\left(\lambda^i(u), \psi\left(\{\!\!\{\lambda^{i}(v)\}\!\!\}\right)\right)$. \cref{eqn:lemma_4_6} becomes 
\begin{equation}
    \lambda^{i+1}(u) := \rho\left(\lambda^i(u), \{\!\!\{\kappa(\lambda^i(v))|(u,v) \in E\}\!\!\}\right),
\end{equation}
which is exactly the vertex colouring function of 1-WL. The proof is done.
\end{proof}

\dfconewl*
\begin{proof}
    Consider two vertices $v$ and $u$, and their 1-hop neighbour vertex sets $N_1(v)$ and $N_1(u)$. We use $\lambda_{\text{dfc}}(\cdot)$ and $\lambda_{\text{wl}}(\cdot)$ to denote the colour refinements by DFC-1 and 1-WL, respectively.
    We first show that if $\lambda_{\text{dfc}}(v) = \lambda_{\text{dfc}}(u)$, then $\lambda_{\text{wl}}(v) = \lambda_{\text{wl}}(u)$. For $\lambda_{\text{dfc}}(v) = \lambda_{\text{dfc}}(u)$ to hold, the number of vertices in $N_1(v)$ must be the same as $N_1(u)$ because of \cref{lemma:equal_neighbour_size}. This means that $v$ and $u$ have the same degree, and thus $\lambda_{\text{wl}}(v) = \lambda_{\text{wl}}(u)$ must hold.
    
    We now show that if $\lambda_{\text{wl}}(v) = \lambda_{\text{wl}}(u)$, $\lambda_{\text{dfc}}(v) = \lambda_{\text{dfc}}(u)$ may not hold. Consider the left graph pair in \cref{fig:circle_examples}, each vertex has a degree of 2 and all vertices have the same colour under 1-WL. However, the number of vertices returned by \cref{eqn:sigma_df} is not the same for the vertices in the left graph and the vertices in the right graph. Specifically, $\eta_{\text{dfc}}(u,E_{\text{tree}}^{T_v},E_{\text{back}}^{T_v})$ returns a set of 3 vertices for each vertex in the left graph and a set of 2 vertices for the right graph. Thus, $\lambda_{\text{dfc}}(v) \neq \lambda_{\text{dfc}}(u)$. This means that DFC-1 can distinguish some pairs of non-isomorphic graphs which 1-WL cannot distinguish. The proof is done.
\end{proof}

\bfcdistinguish*
\begin{proof}
The left graph pair in \cref{fig:circle_examples} can be distinguished by BFC-2 but not by 1-WL. According to \cref{thm:bfcexpressitybeyond}, for any $\delta>2$, BFC-$\delta$ can also distinguish this graph pair. Similarly, the right graph pair in \cref{fig:circle_examples} can be distinguished by BFC-3 but not by 1-WL. So for any $\delta>3$, BFC-$\delta$ can also distinguish this graph pair.
\end{proof}

\dfcdistinguish*
\begin{proof}
The left graph pair in \cref{fig:circle_examples} can be distinguished by DFC-$\delta$ for any $\delta\geq 1$.
The right graph pair in \cref{fig:circle_examples} can be distinguished by DFC-$\delta$ for any $\delta\geq 2$. Both graph pairs cannot be distinguished by 1-WL.
\end{proof}

Before proving \cref{thm:bfc3wl}, we first define the version of the $k$-WL test studied by~\citet{cai1992optimal}, which is also called \emph{folklore-WL} (FWL)~\cite{morris2019weisfeiler}. Let $\overrightarrow{v}=(v_1, v_2,v_3,\dots, v_k)$ be a $k$-tuple. Then the neighbourhood of $k$-FWL is defined as a set of $n=|V|$ elements:

\begin{equation*}
    \hspace*{-3cm}N^F(\overrightarrow{v}) = \{N^F_w(\overrightarrow{v})|w\in V\},
\end{equation*} 
where each $N^F_w(\overrightarrow{v})$ is defined as:
\begin{equation*}
    \hspace{0.6cm}N^F_w(\overrightarrow{v}) = \{(w, v_2,v_3,\dots, v_k), (v_1,w,v_3,\dots, v_k), \dots, (v_1,v_2,\dots, v_{k-1},w) \}.
\end{equation*}

It is known that $k$-WL is equivalent to ($k$-1)-FWL in distinguishing non-isomorphic graphs when $k >2$~\cite{cai1992optimal}. Thus, below we only need to show that the expressivity of BFC is strictly upper bounded by 2-FWL.

When $k=2$, the neighbourhood of $2$-FWL is a set of $n$ elements of a pair of 2-tuples:

\begin{equation}
    \hspace*{-3cm}N^F(u,v) = \{(u,w),(w,v)|w\in V\}.
\end{equation}

Equivalently, the above equation may be written as:

\begin{equation}\label{equ:2-fwl}
    \hspace*{-3cm}N^F(u,v) = \{(u,w,v)|w\in V\}. 
\end{equation}

Let $G=(V,E)$ be an input graph, $2$-FWL assigns colours to all pairs of vertices of $G$. Initially, there are three colours being assigned: \emph{edge}, \emph{nonedge}, and \emph{self}.
Then the colours of these pairs of vertices are refined iteratively by assigning a new colour to each pair $(u, v)$ depending on the colours of $\{(u,w,v)|w\in V\}$ in $G$. This process continues until the colours of all pairs of vertices stabilize.     

Let $Col(u,v)$ denote the colour of the vertex pair $(u,v)$ by 2-FWL, and $d(u,v)$ denote the shortest-path distance between $u$ and $v$. We have \cref{lemma:bfc2fwl}.

\begin{restatable}[]{lemma}{bfc2fwl}
\label{lemma:bfc2fwl}
Given two pairs of vertices $(u,v)$ and $(u',v')$, if $SPG(u,v)\not\simeq SPG(u',v')$, then $Col(u,v)\neq Col(u',v')$.
\end{restatable}

\begin{proof}
We prove this by induction.

\begin{itemize}
    \item When $d(u,v)=0$ and $d(u',v')=0$, we must have $u=v$ and $u'=v'$. Accordingly, both $SPG(u,v)$ and $SPG(u',v')$ contain only one node. Thus $SPG(u,v)\simeq SPG(u',v')$ and it is impossible to have $SPG(u,v)\not\simeq SPG(u',v')$.

\item When $d(u,v)=1$ and $d(u',v')=1$, we have $(u,v)\in E$ and $(u',v')\in E$. Then both $SPG(u,v)$ and $SPG(u',v')$ contain only one edge. Thus $SPG(u,v)\simeq SPG(u',v')$ and it is impossible to have $SPG(u,v)\not\simeq SPG(u',v')$.

\item When $d(u,v)=2$ and $d(u',v')=2$, we must have $(u,v)\not\in E$ and $(u',v')\not\in E$. By Equation~\ref{equ:2-fwl}, we also know that  
\begin{align*}\label{equ:d2-2-fwl}
      N^F(u,v) = & \{(u,w,v)|(u,w)\in E,(w,v)\in E, w\in V\backslash\{u,v\}\}\cup\\&\{(u,w,v)|(u,w)\in E, (w,v)\not\in E, w\in V\backslash\{u,v\}\}\cup\\   
      &\{(u,w,v)|(u,w)\not\in E, (w,v)\in E, w\in V\backslash\{u,v\}\}\cup\\ 
      &\{(u,w,v)|(u,w)\not\in E, (w,v)\not\in E, w\in V\backslash\{u,v\}\}\cup\\ 
      &\{(u,w,v)|w=u\}\cup\\ 
      &\{(u,w,v)|w=v\} 
\end{align*}
Note that, each of the subsets in the above equation corresponds to a different kind of neighbour in the neighbourhood of $(u,v)$. In this case, equivalently, $SPG(u,v)=(V_{uv}, E_{uv})$ can also be expressed as the subsets of $N^F(u,v)$ that corresponds to three kinds of neighbours in the neighbourhood of $(u,v)$ (Lines 1, 5, and 6 in the above equation):
\begin{align*}
      V_{uv} = &\{u,v\}\cup\{w|(u,w)\in E,(w,v)\in E, w\in V\backslash\{u,v\}\}\\
      E_{uv} = & \{(u,w),(w,v)|(u,w)\in E,(w,v)\in E, w\in V\backslash\{u,v\}\}
\end{align*}
We know that the colouring of 2-FWL preserves injectivity. Thus, if $SPG(u,v)\not\simeq SPG(u',v')$, it means that their corresponding subsets in $N^F(u,v)$ and $N^F(u',v')$ are not isomorphic. Then $Col(u,v)\neq Col(u',v')$.

\item Now assume that the statement ``if $SPG(u,v)\not\simeq SPG(u',v')$, then $Col(u,v)\neq Col(u',v')$" holds for any two pairs of vertices $(u,v)$ and $(u',v')$ when $d(u,v)=d(u',v')\leq \Delta$. We want to show that this statement will hold for the case $d(u,v)=d(u',v')= \Delta+1$.

When $d(u,v)=\Delta+1$, we may express $SPG(u,v)$ as a tree rooted at vertex $u$, which has a number of children $\{SPG(u_1,v),\dots, SPG(u_q,v)\}$ where $d(u_i,v)=\Delta$ for $1\leq i\leq q$. Accordingly, we may express $SPG(u',v')$ in a similar way. Thus, if $SPG(u,v)\not\simeq SPG(u',v')$, there are two cases:
\begin{itemize}
    \item[(1)] $\{SPG(u,u),SPG(v,v)\}\not\simeq \{SPG(u',u'),SPG(v',v')\}$ \\By our assumption for the case $d(u,v)=d(u',v')\leq \Delta$, we have $\{Col(u,u),Col(v,v)\}\not\simeq \{Col(u',u'),Col(v',v')\}$. Hence, we know that $Col(u,v)\neq Col(u',v')$.
    \item[(2)] $\{SPG(u,u), SPG(v,v)\}\simeq \{SPG(u',u'),SPG(v',v')\}$\\ Without loss of generality, we assume that $SPG(u,u)\simeq SPG(u',u')$ and $SPG(v,v)\simeq SPG(v',v')$. Then in this case $SPG(u,v)\not\simeq SPG(u',v')$ implies that $\{SPG(u_1,v),\dots, SPG(u_q,v)\}\not\simeq \{SPG(u'_1,v'),\dots, SPG(u'_p,v')\}$ where $(u,u_i)\in E$ and $d(u_i,v)=\Delta$ for $1\leq i\leq q$, and $(u',u'_j)\in E$ and $d(u'_j,v')=\Delta$ for $1\leq j\leq p$. Here, $p=q$ must hold; otherwise we immediately have $Col(u,v)\neq Col(u',v')$. By our assumption for the case $d(u,v)=d(u',v')\leq \Delta$, we have $\{Col(u_1,v),\dots, Col(u_q,v)\}\not\simeq \{Col(u'_1,v'),\dots, Col(u'_p,v')\}$. Thus, $Col(u,v)\neq Col(u',v')$ must hold.
\end{itemize}
\end{itemize}
\end{proof}

\bfcthreewl*
\begin{proof}
We first seek to show 3-WL is at least as powerful as BFC-$\delta$. Let $G=(V_G, E_G)$ and $H=(V_H, E_H)$ be two input graphs, according to \cref{lemma:lvcbfs_esgp}, BFC can distinguish graphs only when they have different ESPGs, e.g. $\{\!\!\{\lambda_{S_v}(v): v\in V_G\}\!\!\} \neq \{\!\!\{\lambda_{S_u}(u): u\in V_H\}\!\!\}$. So we just need to show $\{\!\!\{Col(v,v') | v' \in V_G \}\!\!\} \neq \{\!\!\{Col(u,u') | u' \in V_H \}\!\!\}$ for any $v\in V_G$ and $u\in V_H$ when $S_v \not\simeq S_u$. 
$S_v\not\simeq S_u$ implies $\{\!\!\{SPG(v', v): v'\in V_G\}\!\!\} \not\simeq \{\!\!\{SPG(u', u): u\in V_H\}\!\!\}$. According to \cref{lemma:bfc2fwl}, we must have $\{\!\!\{Col(v,v') | v' \in V_G \}\!\!\} \neq \{\!\!\{Col(u,u') | u' \in V_H \}\!\!\}$. Thus 3-WL can distinguish graphs when they have different ESPGs, so 3-WL is at least as powerful as BFC-$\delta$.

Now we seek to show the strictness of this bound, that is 3-WL is strictly more expressive than BFC. We show this by introducing an example graph pair in \cref{fig:bfc3wl}. In this example 3-WL can distinguish the graphs but BFC cannot because the two graphs have isomorphic ESPGs. Therefore the expressiveness of BFC is strictly upper bounded by 3-WL.

\begin{figure*}[ht]
\centering
\includegraphics[clip,width=0.4\textwidth]{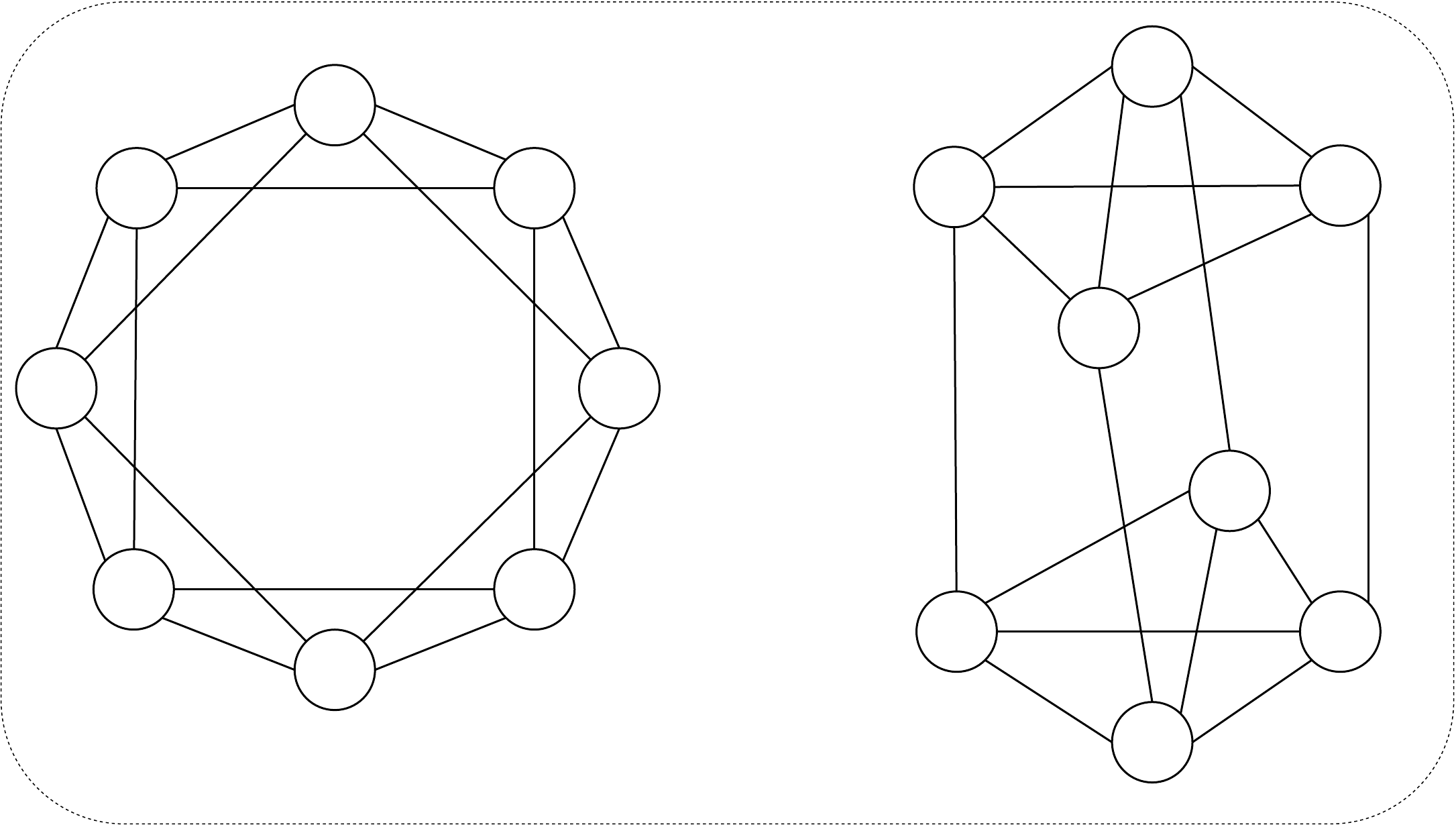}
\caption{A pair of non-isomorphic graphs that can be distinguished by 3-WL but not by BFC.}
\label{fig:bfc3wl}
\end{figure*}

\end{proof}

To prove \cref{thm:bfcexpressitybeyond}, we first introduce \cref{lemma:bfc_equal}.



\begin{restatable}[]{lemma}{expressityequal}
\label{lemma:bfc_equal}
BFC-${\delta+1}$ is at least as expressive as BFC-${\delta}$ in distinguishing non-isomorphic graphs.
\end{restatable}

\begin{proof}
\label{proof:as_express_as}
This lemma requires to show that, for any $i$-th iteration, if BFC-$\delta+1$ must have the same multiset of vertex colours for $G_1$ and $G_2$, then BFC-$\delta$ must also have the same multisets of vertex colours for $G_1$ and $G_2$.
Below we show for any iteration $i$, if the colours of any two vertices in $G_1$ and $G_2$ are the same by BFC-$\delta+1$, then their colours by BFC-$\delta$ must also be the same.
We show this by induction.
 
\begin{itemize}
    \item For $i=0$, it is obvious that the initial vertex colours are the same for BFC-$\delta+1$ and BFC-$\delta$.
    \item For $i>0$, we assume that this statement ``if $\lambda^{i}(u_1) = \lambda^{i}(u_2)$ for BFC-$\delta+1$ then $\lambda^{i}(u_1) = \lambda^{i}(u_2)$ for BFC-$\delta$" holds for the $i$-th iteration, and seek to show that the statement also holds for the $(i+1)$-th iteration. We show this by contradiction. Assuming $\lambda^{i+1}(u_1) = \lambda^{i+1}(u_2)$ hold for BFC-$\delta+1$ but not for BFC-$\delta$, we have
    \begin{equation}
    \label{eqn:expand_delta_plus_1}
        \rho(\lambda^{i}(u_1), \{\!\!\{\lambda^{i}_{v}(u_1)| v\in N_{\delta+1}(u_1)\}\!\!\}) 
        =  \rho(\lambda^{i}(u_2), \{\!\!\{\lambda^{i}_{v}(u_2)| v\in N_{\delta+1}(u_2)\}\!\!\})
    \end{equation}
    and 
    \begin{equation}
    \label{eqn:expand_delta}
        \rho(\lambda^{i}(u_1), \{\!\!\{\lambda^{i}_{v}(u_1)| v\in N_{\delta}(u_1)\}\!\!\}) 
        \neq  \rho(\lambda^{i}(u_2), \{\!\!\{\lambda^{i}_{v}(u_2)| v\in N_{\delta}(u_2)\}\!\!\}).
    \end{equation}
    Because $\lambda^{i+1}(u_1) = \lambda^{i+1}(u_2)$ for BFC-$\delta+1$, we must have $\lambda^{i}(u_1) = \lambda^{i}(u_2)$ for BFC-$\delta+1$. According to our assumption ``if $\lambda^{i}(u_1) = \lambda^{i}(u_2)$ for BFC-$\delta+1$ then $\lambda^{i}(u_1) = \lambda^{i}(u_2)$ for BFC-$\delta$", we have $\lambda^{i}(u_1) = \lambda^{i}(u_2)$ hold for BFC-$\delta$. Thus, we can simplify \cref{eqn:expand_delta_plus_1,eqn:expand_delta} as 
    \begin{equation}
        \label{eqn:color_multiset_delta_plus_1}
        \{\!\!\{\lambda^{i}_{v}(u_1)| v\in N_{\delta+1}(u_1)\}\!\!\}  =  \{\!\!\{\lambda^{i}_{v}(u_2)| v\in N_{\delta+1}(u_2)\}\!\!\}
    \end{equation}
    \begin{equation}
        \label{eqn:color_multiset_delta}
        \{\!\!\{\lambda^{i}_{v}(u_1)| v\in N_{\delta}(u_1)\}\!\!\}  \neq  \{\!\!\{\lambda^{i}_{v}(u_2)| v\in N_{\delta}(u_2)\}\!\!\}
    \end{equation}
    Because $N_{\delta}(u) \subseteq N_{\delta+1}(u)$ and $N_{\delta+1}(u) - N_{\delta}(u) = \{v|d(u,v)=\delta+1\}$, 
    there must exist at least one pair of vertices $u'$ and $u''$, where $d(v,u')=\delta+1$ and $d(v,u'')\leq\delta$, such that $\lambda^{i}_v(u')=\lambda^{i}_v(u'')$.
    This contradicts \cref{lemma:diff_lambda_diff_hop}. So the assumption ``$\lambda^{i+1}(u_1) \neq \lambda^{i+1}(u_2)$ for BFC-$\delta$" must not hold. Thus, the statement ``if $\lambda^{i+1}(u_1) = \lambda^{i+1}(u_2)$ for BFC-$\delta+1$ then $\lambda^{i+1}(u_1) = \lambda^{i+1}(u_2)$ for BFC-$\delta$" holds.
\end{itemize}
This means that, for any iteration $i$, if the colours of any two vertices in $G_1$ and $G_2$ are the same by BFC-$\delta+1$, then their colours by BFC-$\delta$ must also be the same.
The proof is done.
\end{proof}

Now we prove \cref{thm:bfcexpressitybeyond}.
\bfcexpressitybeyond*
\begin{proof}
By \cref{lemma:bfc_equal}, we know that BFC-$\delta+1$ is at least as expressive as BFC-$\delta$. Now we just need to show that there exists at least one pair of non-isomorphic graphs $(\hat{G}_1, \hat{G}_2)$ that can be distinguished by BFC-$\delta+1$ but not by BFC-$\delta$.

Inspired by \citet{wang2023mathscrnwl}, we hereby show a specific construction of such graph pairs using cycles. We construct $\hat{G}_1$ to be two cycles of length $2\delta+1$, and $\hat{G}_2$ to be one cycle of length $4\delta+2$, for any $\delta\geq 1$. $\hat{G}_1$ and $\hat{G}_2$ can be distinguished by BFC-$\delta+1$ but not by BFC-$\delta$. Figure \ref{fig:circle_examples} shows two examples graph pairs constructed using this method. The proof is done.
\end{proof}




\lvcgnn*

\begin{proof}
    The concatenation operator in Equation \ref{eqn:gnn_layer} preserves injectivity. So Equation \ref{eqn:gnn_layer} is equivalent of replacing $\rho(\cdot)$ in Equation \ref{eqn:lvc-all} with an MLP. The summation operator in Equation \ref{eqn:gnn_huv} is the injective variant of $\psi(\cdot)$ in Equation \ref{eqn:lvc}. The multiplication with $W_c$ is a single-layer MLP that used in place for $\phi(\cdot)$ in Equation \ref{eqn:lvc}. As a universal approximator~\citep{DBLP:journals/nn/Hornik91,DBLP:journals/nn/HornikSW89}, MLP can learn injective functions.
    Hence, \model{} can be shown to be as expressive as LVC following the proof of Theorem 3 by \citet{xu2018powerful}. We leave the details out for brevity.
\end{proof}